\documentclass[final,5p,times,twocolumn]{elsarticle}

\usepackage{hyperref}
\usepackage{pgfplots}
\usepackage{newfloat}
\usepackage{listings}
\usepackage{algorithm}
\usepackage{alltt}
\usepackage{upquote}
\usepackage{booktabs}

\usepackage{amssymb,amsmath}
\usepackage{bm}
\usepackage{color}
\usepackage{xcolor}
\usepackage{dcolumn}
\usepackage{graphicx}
\usepackage{multirow}
\usepackage{subfigure}
\usepackage{kotex}
\usepackage{enumitem}
\usepackage{lineno}
\usepackage{adjustbox}
\usepackage{pifont}
\newcommand{\cmark}{\ding{51}} 
\newcommand{\xmark}{\ding{55}} 

\journal{Expert Systems with Applications}

\begin{document}

\begin{frontmatter}



\title{FiGKD: Fine-Grained Knowledge Distillation via High-Frequency Detail Transfer}


\author[1]{Seonghak Kim} 
\ead{seonghak35@gmail.com}

\affiliation[1]{organization={Agency for Defense Development (ADD)},
            country={Republic of Korea}}


\tnotetext[]{\copyright~2026 This is the author’s accepted manuscript of an article published in Expert Systems with Applications, made available under the CC-BY-NC-ND 4.0 license: https://creativecommons.org/licenses/by-nc-nd/4.0/ \\ Digital Object Identifier (DOI): 10.1016/j.eswa.2026.132071}

\begin{abstract}
Knowledge distillation (KD) is a widely adopted technique for transferring knowledge from a high-capacity teacher model to a smaller student model by aligning their output distributions. However, existing methods often underperform in fine-grained visual recognition tasks, where distinguishing subtle differences between visually similar classes is essential. This performance gap stems from the fact that conventional approaches treat the teacher’s output logits as a single, undifferentiated signal—assuming all contained information is equally beneficial to the student. Consequently, student models may become overloaded with redundant signals and fail to capture the teacher’s nuanced decision boundaries. To address this issue, we propose Fine-Grained Knowledge Distillation (FiGKD), a novel frequency-aware framework that decomposes a model’s logits into low-frequency (content) and high-frequency (detail) components using the discrete wavelet transform (DWT). FiGKD selectively transfers only the high-frequency components, which encode the teacher’s semantic decision patterns, while discarding redundant low-frequency content already conveyed through ground-truth supervision. Our approach is simple, architecture-agnostic, and requires no access to intermediate feature maps. Extensive experiments on CIFAR-100, TinyImageNet, and multiple fine-grained recognition benchmarks show that FiGKD consistently outperforms state-of-the-art logit-based and feature-based distillation methods across a variety of teacher–student configurations. These findings confirm that frequency-aware logit decomposition enables more efficient and effective knowledge transfer, particularly in resource-constrained settings.
\end{abstract}



\end{frontmatter}



\section{Introduction}
\label{sec:introduction}
Deep learning has driven remarkable progress in computer vision, achieving state-of-the-art performance in tasks such as image classification~\cite{resnet,vgg, vision_trans}, object detection~\cite{yolo, rcnn}, semantic segmentation~\cite{fcn, pyramid, deeplab}, panoptic perception~\cite{yolop, yolopv2}, and multi-modal fusion~\cite{multi, multimodal}. These advancements have been largely enabled by the increasing scale of deep neural networks, including architectures with billions of parameters and deeply layered hierarchies. While such overparameterized models deliver exceptional accuracy, their substantial computational and memory requirements pose significant challenges for deployment in resource-constrained environments, such as edge devices, real-time systems, and embedded platforms used in industrial and automotive applications. To address these limitations, a variety of model compression techniques have been proposed, including pruning~\cite{pruning}, quantization~\cite{quantization}, low-rank factorization~\cite{lowrank}, and knowledge distillation (KD)~\cite{kd_survey}. Among these, KD has emerged as a particularly promising approach due to its simplicity, general applicability, and strong effectiveness. In KD, a compact student model is trained to imitate the behavior of a larger teacher model, typically by aligning their output logits or internal feature representations. This enables the student to approach—and in some cases even surpass—the performance of the teacher, despite having far fewer parameters.

\begin{figure}[t]
  \centering
  \includegraphics[width=1.0\columnwidth]{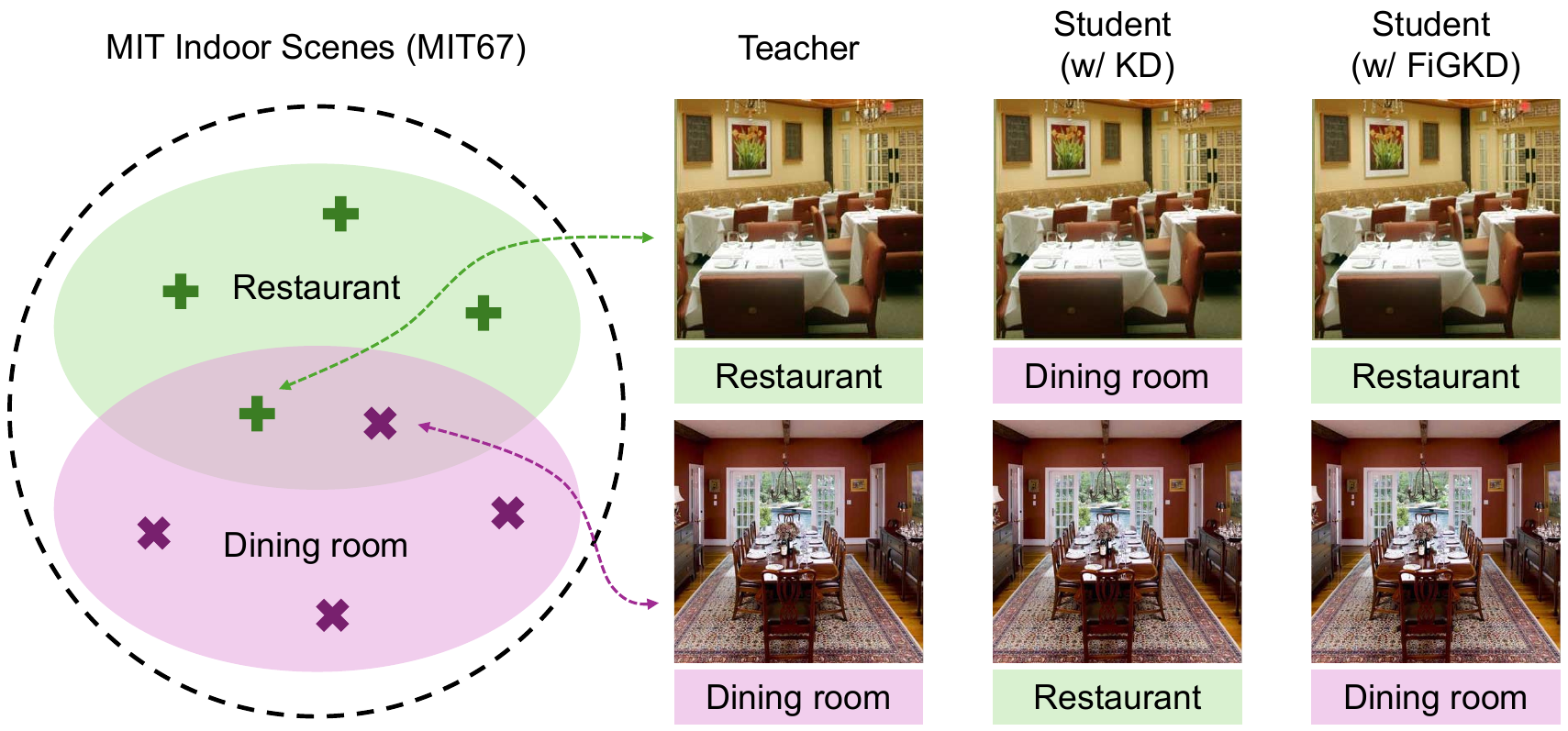}
  \caption{Comparison of teacher, conventional KD, and FiGKD models on a fine-grained classification task.}
  \label{fig:comparison}
\end{figure}

Since its introduction by Hinton~\cite{hinton}, knowledge distillation (KD) has been broadly categorized into logit-based~\cite{hinton, dkd} and feature-based~\cite{fitnet, review} approaches. Logit-based distillation supervises the student using the teacher’s final output logits, whereas feature-based methods transfer intermediate representations. Although feature-based KD often achieves superior performance due to the rich spatial and semantic information captured in intermediate features, it also introduces several practical challenges. First, feature-based methods typically require tight architectural coupling between the teacher and student networks, which limits flexibility in model design. Second, transferring intermediate feature maps incurs significant memory and bandwidth overhead, especially in large-scale models or real-time applications. Finally, in federated learning or privacy-sensitive environments, accessing internal representations may pose serious security risks, as these activations could potentially leak sensitive information about the input data. As a result, logit-based distillation is often preferred in real-world deployment scenarios due to its simplicity, efficiency, and model-agnostic nature — requiring only access to the final output logits~\cite{mlkd, r2kd, energykd, tssk}.

However, traditional logit-based knowledge distillation (KD) methods remain limited in their effectiveness for fine-grained visual recognition tasks, which demand capturing subtle semantic differences between classes. As illustrated in Fig.~\ref{fig:comparison}, student models trained with conventional KD frequently struggle to distinguish visually similar categories, in contrast to their teacher counterparts. This limitation occurs because traditional logit-based KD treats all information within the teacher’s logits as equally valuable, ignoring the fact that logits encode both coarse content (e.g., dominant class scores) and fine-grained semantic detail (e.g., distributional patterns over non-target classes). This entanglement prevents the student from selectively focusing on the most informative parts of the teacher’s knowledge — especially when the teacher is highly overparameterized and its output includes redundant or overconfident elements that the student cannot effectively replicate.

\begin{figure}[tbp]
  \centering
  \includegraphics[width=1.0\columnwidth]{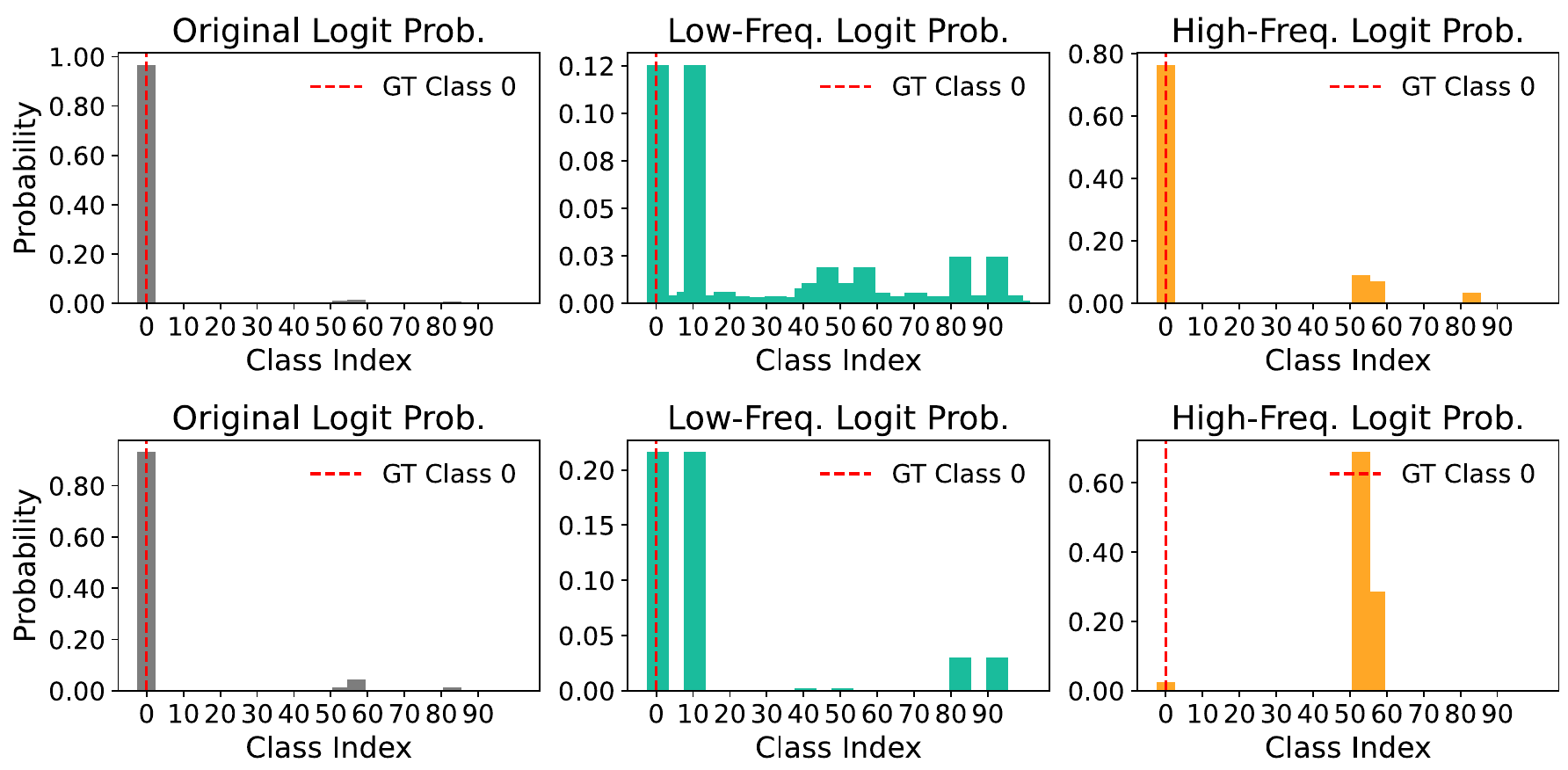}
  \caption{Visualization of probability distributions across frequency components for ResNet32x4 (Top) and ResNet8x4 (Bottom) on CIFAR-100.}
  \label{fig:logit_analysis}
\end{figure}

To better understand the limitations of full-logit distillation, we conducted an empirical analysis of the frequency components of the logits. Specifically, we applied a 2D discrete wavelet transform (DWT) to the logits produced by both a high-capacity model and a smaller model. As shown in Fig.~\ref{fig:logit_analysis}, we observed that strong, high-accuracy models (e.g., ResNet32x4) tend to preserve class-discriminative information even in the high-frequency components alone — they can correctly predict the target class using only the high-frequency logits. In contrast, lower-capacity models (e.g., ResNet8x4) are typically unable to do so: while they may make correct predictions when using the full logits, they frequently fail when relying solely on the high-frequency part. Additional visualizations are provided in Sec.~\ref{subsec:visualizations}. These findings suggest that high-performing models encode meaningful semantic structure in the high-frequency components, whereas weaker models struggle to capture such detail. This observation motivates our approach: rather than transferring the entire logit vector, we propose to distill only the teacher’s high-frequency detail information, which captures the subtle decision patterns that the student would otherwise miss.

Consequently, to overcome the limitations of conventional logit-based distillation in handling subtle classification tasks, we introduce Fine-Grained Knowledge Distillation (FiGKD) — a lightweight yet effective framework that focuses on transferring the most informative part of the teacher's output. Rather than matching the entire logit vector, FiGKD identifies and distills the teacher’s high-frequency components, which are shown to encode the teacher’s fine-grained decision behavior and semantic differentiation between closely related classes. This is achieved through a simple process: we apply a 2D discrete wavelet transform (DWT) to the teacher’s logits, separating them into low-frequency content and high-frequency detail signals. FiGKD leverages this decomposition by selectively guiding the student to imitate only the detail — that is, the fine-grained structure within the teacher’s predictions — while relying on the ground-truth labels to supervise the overall class identity. This targeted supervision helps the student allocate its limited capacity more effectively, promoting generalization in tasks that require detailed semantic understanding.

We evaluate FiGKD across diverse benchmarks, including CIFAR-100, TinyImageNet, and multiple fine-grained visual recognition (FGVR) datasets. The results demonstrate that FiGKD consistently improves student performance over state-of-the-art logit-based and feature-based KD methods, confirming that detail-focused distillation in the frequency domain offers a powerful and transferable learning signal.

\section{Related Works}
\label{sec:related_works}

\subsection{Knowledge Distillation}
\label{subsec:knolwedge_distillation}

Knowledge Distillation (KD) was first introduced by Hinton~\cite{hinton} as a model compression technique, where a smaller student network is trained to mimic the softened output probabilities of a larger teacher network. These softened logits—obtained by applying a higher temperature to the softmax function—reveal how the teacher generalizes across classes, thereby providing richer supervision than hard labels alone. Since its introduction, KD has been widely adopted and extended. Most approaches combine the distillation loss with the standard cross-entropy loss on ground-truth labels, forming a joint objective that balances learning from both the teacher and the data.

Existing KD methods are generally categorized into logit-based~\cite{hinton, dkd} and feature-based~\cite{fitnet, review} approaches. {However, recent advancements have further diversified the distillation paradigm beyond these standard categories. For instance, FFKD~\cite{ffkd} introduces a reciprocal distillation framework that alternates between forward (teacher to student) and feedback (student to teacher) knowledge distillation, allowing the teacher to adjust its teaching strategy dynamically. Additionally, IPASD~\cite{ipasd} explores self-distillation by leveraging intra-class progressive mechanisms to generate soft labels from the model itself, thereby reducing reliance on a fixed external teacher.}

{Focusing on logit-based methods, these approaches operate directly on model outputs—either logits or predicted probabilities.} Hinton's original formulation falls under this category, along with numerous subsequent variants that refine how the teacher’s output distribution is leveraged. For example, Decoupled Knowledge Distillation (DKD)~\cite{dkd} explicitly splits the logit vector into two parts: the target class and the non-target classes. By assigning separate weights to each component, DKD enhances supervision for non-target predictions, based on the insight that high-confidence predictions can suppress meaningful inter-class relationships. This design improves the distillation process by emphasizing the relative importance of secondary class scores.

FiGKD shares a similar motivation—focusing on the fine-grained structure within the logit vector—but takes a fundamentally different approach. Rather than explicitly separating logits into target and non-target classes, FiGKD decomposes the entire logit vector in the frequency domain into high- and low-frequency components. This enables the model to capture fine-grained class-wise contrast—often embedded in high-frequency variations—without relying on explicit target-class supervision. As a result, FiGKD naturally preserves inter-class structural information that is often overlooked in traditional full-logit distillation.

\subsection{Feature Distillation}
\label{subsec:feature_distillation}

Feature-based knowledge distillation (KD) methods aim to transfer rich internal representations from the teacher to the student. As an early attempt, FitNets~\cite{fitnet} introduced the concept of hint layers, where the student learns to replicate intermediate feature maps from the teacher network. Since then, numerous variants have emerged to capture different structural properties of the feature space. Attention Transfer (AT)~\cite{at}, for instance, guides the student to mimic the spatial attention maps of the teacher, highlighting where the teacher focuses within the input. Relational Knowledge Distillation (RKD)~\cite{rkd} encourages the student to preserve the pairwise geometric relationships between samples found in the teacher’s feature space. Probabilistic Knowledge Transfer (PKT)~\cite{pkt} aligns the overall feature distribution between the teacher and student by matching their probability distributions rather than raw feature values. Contrastive Representation Distillation (CRD)~\cite{crd} adopts a contrastive learning framework, training the student to bring its representations closer to those of the teacher using positive and negative sample pairs. 

Among more recent approaches, Review KD~\cite{review} introduces a review mechanism that summarizes multi-level knowledge from the teacher and guides the student using these condensed representations. DiffKD~\cite{diffkd} treats knowledge distillation as a denoising process, where a teacher-guided diffusion model refines the student's noisy features into cleaner, more informative representations. CAT-KD~\cite{catkd} focuses on class-discriminative regions by transferring class activation maps (CAMs), enabling the student to attend to the same spatial regions as the teacher for each class. While these methods demonstrate strong performance by modeling richer intermediate representations, they often require tight architectural coupling, additional modules, or direct access to internal features—factors that complicate deployment in practical systems.

In contrast, FiGKD distills fine-grained knowledge solely from the final logits. By decomposing these logits into frequency components and selectively transferring high-frequency details, FiGKD captures the semantic subtlety of the teacher’s predictions without accessing on internal features. This makes it lightweight, architecture-agnostic, and particularly well-suited for fine-grained visual recognition tasks.

\subsection{Frequency Domain}
\label{subsec:frequency_domain}

Recent research has increasingly explored the role of frequency-domain representations in neural networks~\cite{freq_learning1, freq_learning2, freq_learning3, freq_learning4}. This growing interest stems from the observation that neural networks exhibit varying sensitivities across frequency bands~\cite{freq_learning}: low-frequency components typically encode coarse global structures, while high-frequency components capture fine-grained details. In image generation, frequency-based approaches have been employed to improve output fidelity. For instance, wavelet-based image translation methods~\cite{wavelet_trans} apply the discrete wavelet transform (DWT) to decompose generated images into multi-scale frequency bands. By emphasizing or supervising the high-frequency components, these methods enable lightweight generators to better reconstruct textures and sharp details. Such findings suggest that high-frequency information is particularly challenging for compact models to learn, yet critical for producing perceptually realistic outputs.

Beyond generation tasks, other studies have investigated the frequency sensitivity of convolutional neural networks (CNNs) themselves. Fourier-based analyses of model robustness have revealed that CNNs tend to favor low-frequency features~\cite{freq_learning}, and that robustness to noise and perturbations is closely linked to a model’s ability to handle high-frequency signals~\cite{fourier_robust}. These insights further support the use of frequency-domain analysis as a tool to understand model behavior and generalization. 

In the context of cross-modal knowledge transfer, frequency decomposition has been used to disentangle modality-specific and modality-agnostic representations: low-frequency components often encode shared content across modalities, while high-frequency components capture modality-specific variations~\cite{freq_modal}. Applying different loss functions to each component has been shown to improve generalization across domains such as audio and vision.

Most relevant to our work, frequency-domain representations have recently been explored for fine-grained recognition. For example, FicNet~\cite{ficnet} introduces a Multi-Frequency Neighborhood (MFN) module that combines neighborhood modeling with discrete cosine transform (DCT) to capture discriminative structural patterns for few-shot fine-grained classification.

While these prior works demonstrate the value of frequency-domain analysis, our method differs fundamentally in both application domain and technical approach.
Specifically, although FicNet demonstrates the effectiveness of frequency components for fine-grained recognition in a metric-based few-shot setting, our work targets knowledge distillation under standard supervised learning.
Also, in contrast to image generation methods that apply DWT to the image space to enhance pixel-level fidelity, we apply DWT directly to the logit vector of neural networks. {Whereas prior work focuses on visual reconstruction or feature metric alignment}, our goal is to improve semantic discriminability in the output space. Specifically, we treat the logit vector as a structured signal and decompose it into low- and high-frequency components. We show that the high-frequency component captures subtle inter-class distinctions that full-logit matching often misses, providing an effective supervisory signal for training student models. This enables more effective knowledge transfer of fine-grained semantic patterns without requiring access to intermediate feature maps, and proves especially beneficial in fine-grained visual recognition tasks where decision boundary precision is essential.

\section{Methods}
\label{sec:methods}

\begin{figure*}[htbp]
  \centering
  \includegraphics[width=0.8\linewidth]{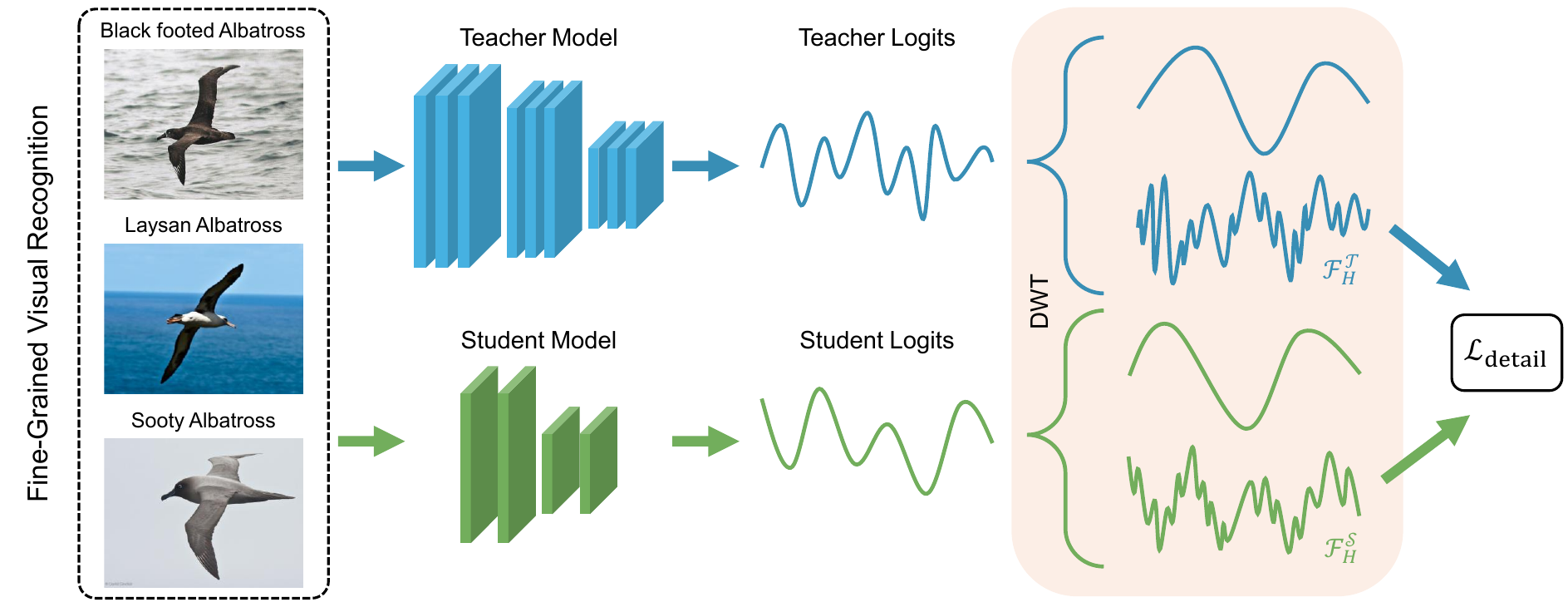}
  \caption{Overview of Fine-Grained Knowledge Distillation framework. Given an input image, the teacher and student produce their logits, which are reshaped and decomposed via a 2D DWT to obtain the high-frequency components ($\mathcal{F}^\mathcal{T}_H$ and $\mathcal{F}^\mathcal{S}_H$). The detail loss ($\mathcal{L}_\text{detail}$) is applied to align these high-frequency components.}
  \label{fig:fine_grained_results}
\end{figure*}

In this section, we present Fine-Grained Knowledge Distillation (FiGKD), a frequency-aware distillation framework designed to improve fine-grained recognition by selectively transferring high-frequency components of the teacher's output logits. We begin by revisiting conventional logit-based distillation methods and their limitations, then introduce the discrete wavelet transform (DWT), and finally detail the formulation of FiGKD.

\subsection{Revisiting Logit Distillation}
\label{subsec:revisit_logit}

Knowledge distillation (KD) is a widely used paradigm for transferring knowledge from a high-capacity teacher network to a lightweight student network~\cite{hinton}. In its most common form, KD encourages the student to mimic the softened output probabilities of the teacher, thereby transferring so-called \textit{dark knowledge}~\cite{energykd, r2kd}—information about class similarities encoded in the teacher's output distribution.

Given an input sample $x$ with ground-truth label $y$, $\ell^{\mathcal{T}}(x) \in \mathbb{R}^C$ and $\ell^{\mathcal{S}}(x) \in \mathbb{R}^C$ are denoted by the pre-softmax logits produced by the teacher $\mathcal{T}$ and student $\mathcal{S}$ networks, respectively, where $C$ is the number of classes. In the classic logit-based KD, the student is trained to minimize a weighted sum of the cross-entropy loss $(\mathcal{L}_\textbf{CE})$ with ground-truth and a distillation loss based on the Kullback-Leibler (KL) divergence $(\mathcal{D}_\text{KL})$ between the softened output distributions:
\begin{equation}
\mathcal{L}_\text{KD} = (1 - \lambda) \mathcal{L}_\text{CE}(\boldsymbol{p}^{\mathcal{S}}, y) + \lambda T^2 \, \mathcal{D}_\text{KL}(\boldsymbol{q}^{\mathcal{T}} \, \| \, \boldsymbol{q}^{\mathcal{S}}),
\end{equation}
where $\boldsymbol{p}^{\mathcal{S}} = \mathrm{softmax}(\ell^{\mathcal{S}})$ is the standard output of the student, and $\boldsymbol{q}^{\mathcal{T}} = \mathrm{softmax}(\ell^{\mathcal{T}}/T)$ and $\boldsymbol{q}^{\mathcal{S}} = \mathrm{softmax}(\ell^{\mathcal{S}}/T)$ are the softened teacher and student outputs, respectively, with temperature $T$. The hyperparameter $\lambda$ controls the trade-off between supervised learning and distillation.

Here, $\boldsymbol{p}^{\mathcal{S}} = \mathrm{softmax}(\ell^{\mathcal{S}})$ is the standard output of the student, while $\boldsymbol{q}^{\mathcal{T}} = \mathrm{softmax}(\ell^{\mathcal{T}}/T)$ and $\boldsymbol{q}^{\mathcal{S}} = \mathrm{softmax}(\ell^{\mathcal{S}}/T)$ denote the teacher and student logits softened by a temperature parameter $T$. The hyperparameter $\lambda$ controls the balance between supervised learning and distillation. While vanilla KD enables the transfer of global class relationships, it treats all logit components uniformly and may fail to emphasize the subtle semantic distinctions that are critical in fine-grained recognition. 

A recent advancement, Decoupled Knowledge Distillation (DKD)~\cite{dkd}, attempts to separate the contributions of target and non-target classes in the loss function.
\begin{equation}
\mathcal{L}_\text{DKD} = \alpha \mathcal{L}_\text{TCKD} + \beta \mathcal{L}_\text{NCKD},
\end{equation}
where $\mathcal{L}_\text{TCKD}$ and $\mathcal{L}_\text{NCKD}$ denote the losses computed over the target and non-target class logits, respectively, and $\alpha, \beta$ are balancing coefficients. This formulation further highlights the importance of inter-class relationships among non-target classes. However, despite these refinements, conventional logit-based distillation still blends information across the entire logit vector, without explicitly disentangling dominant class predictions from the nuanced inter-class relationships that underlie fine-grained decision boundaries. Furthermore, because these methods operate directly in the raw logit space, they are unable to capture localized patterns or frequency-specific structures that may reflect subtle  yet semantically meaningful variations among classes.

In the following sections, we address these limitations by introducing a frequency-domain decomposition of logits via the Discrete Wavelet Transform (DWT), enabling more targeted and effective knowledge transfer for fine-grained recognition tasks.

\subsection{Discrete Wavelet Transform}
\label{subsec:wavelet_transform}

To enable frequency-domain decomposition of logit vectors, we leverage the Discrete Wavelet Transform (DWT)~\cite{wavelet}, a classical signal processing technique that provides localized analysis in both spatial and frequency domains. Unlike the Fourier transform, which captures only global frequency content, the DWT provides multi-resolution representations by decomposing a signal into scaled and shifted versions of a base function known as the mother wavelet.

The wavelet basis is constructed by dilating and translating the mother wavelet function 
$\varphi(t)$:

\begin{equation} \varphi_{j, k}(t) = 2^{-j/2} \varphi(2^{-j}t - k), \quad j, k \in \mathbb{Z} \end{equation}
{where} $j$ and $k$ denote the scale and translation parameters, respectively, which control how the mother wavelet is dilated and shifted along the input signal. Both $j$ and $k$ take values in the set of integers $\mathbb{Z}$. When $\varphi$ is an orthogonal wavelet, these functions form an orthonormal basis in $L^2(\mathbb{R})$, enabling compact and non-redundant signal representations.

Given a 2D input signal $x$, the 2D DWT operator, denoted as $\mathcal{W}_\varphi(\cdot)$, decomposes the signal into a low-frequency component \( \mathcal{F}_L(x) \), which captures coarse-scale structure, and a high-frequency component \( \mathcal{F}_H(x) \), which contains fine-grained details. This decomposition effectively separates primary content from subtle variations, facilitating multi-scale analysis and representation.

\subsection{Fine-Grained Knowledge Distillation}
\label{subsec:figkd}

Building upon the frequency decomposition framework, we propose FiGKD—a logit-based distillation method that selectively transfers high-frequency knowledge from teacher to student, as shown in Fig.~\ref{fig:fine_grained_results}. The key intuition behind FiGKD is that fine-grained visual distinctions, which are crucial for differentiating semantically similar classes, are primarily encoded in the high-frequency components of the logit signal. These are often overlooked in conventional knowledge distillation (KD), which treats all logit information equally. 

Given a classification task with $C$ classes, each logit vector $\ell \in \mathbb{R}^C$ is interpreted as a one-dimensional semantic signal. To apply the DWT, we reshape this vector into a matrix $\hat{\ell} \in \mathbb{R}^{H \times W}$ such that $H \cdot W = C$\footnote{If $C$ is not a perfect square, $H$ and $W$ are chosen to be as close as possible to $\sqrt{C}$ and $\lceil C / H \rceil$, respectively, such that $H \cdot W = C$. For example: for TinyImageNet ($C=200$), we use $H=10$, $W=20$; for CUB-200, $H=10$, $W=20$; for Stanford Dogs ($C=120$), $H=10$, $W=12$; for MIT67 ($C=67$), $H=1$, $W=67$; and for Stanford40 ($C=40$), $H=5$, $W=8$.}. We fix the decomposition level to $J=1$ and use the Haar wavelet for efficiency and simplicity.

Although the logit vector is inherently one-dimensional, we adopt a 2D discrete wavelet transform rather than a 1D variant. This design allows the reshaped logits to be decomposed into three orthogonal high-frequency subbands (horizontal, vertical, and diagonal), providing multi-directional structural supervision. As demonstrated in Sec.~\ref{subsec:ablation_study}, this 2D decomposition yields consistently better performance than 1D-DWT by enforcing richer fine-grained constraints on the student.

Given the teacher and student logits $\ell^\mathcal{T}, \ell^\mathcal{S} \in \mathbb{R}^C$, we reshape them to matrices $\hat{\ell}^\mathcal{T}, \hat{\ell}^\mathcal{S} \in \mathbb{R}^{H \times W}$. Applying a single-level 2D DWT yields:

\begin{equation}
\mathcal{F}_L, \mathcal{F}_H = \mathcal{W}_\varphi(\hat{\ell}),
\end{equation}
where $\mathcal{F}_L$ and $\mathcal{F}_H$ represent the low-frequency and high-frequency components of the logits, respectively.

We perform DWT on both teacher and student logits:
\begin{equation}
\mathcal{F}_L^\mathcal{T}, \mathcal{F}_H^\mathcal{T} = \mathcal{W}_\varphi(\hat{\ell}^\mathcal{T}), \quad
\mathcal{F}_L^\mathcal{S}, \mathcal{F}_H^\mathcal{S} = \mathcal{W}_\varphi(\hat{\ell}^\mathcal{S}).
\end{equation}
To distill only the high-frequency semantic structure from the teacher, we define a high-frequency loss computed between corresponding subbands of the teacher and student logits. We refer to this term as a detail loss, as it encourages the student to learn the fine-grained variation patterns encoded in the teacher’s high-frequency logit components. In contrast to losses computed in the pixel or feature space, our detail loss directly compares semantic structure in the logit domain.

Formally, for each sample \( i \in \{1, \dots, B\} \) in a batch of size \( B \), and for each high-frequency subband \( k \in \{1, \dots, K \} \), the distillation loss is defined as:

\begin{equation}
\mathcal{L}_{\text{detail}} = \frac{1}{B} \sum_{i=1}^{B} \sum_{k=1}^{K} \left\| \mathcal{F}_{H_k}^{\mathcal{T}}[i] - \mathcal{F}_{H_k}^{\mathcal{S}}[i] \right\|_2^2
\end{equation}
where $\mathcal{F}_{H_k}^{\mathcal{T}}$ and $\mathcal{F}_{H_k}^{\mathcal{S}}$ denote the $k$-th high-frequency subbands of the teacher and student logits, respectively.

This frequency-aware logit decomposition enables semantic disentanglement: we discard the dominant low-frequency content and instead emphasize high-frequency subbands that encode critical fine-grained decision patterns. This selective focus is particularly beneficial for improving student models in fine-grained recognition tasks, where subtle differences between classes are crucial.

In addition to the detail loss, we keep the standard cross-entropy loss $\mathcal{L}_{\text{CE}}$ for supervised learning. The final training objective is a weighted combination of the two:
\begin{equation}
    \mathcal{L}_\text{FiGKD} = \alpha \mathcal{L}_\text{CE} + \beta \mathcal{L}_\text{detail}
\end{equation}
where $\alpha$ and $\beta$ are balancing hyperparameters.

Unlike previous approaches that apply wavelet transforms to raw images~\cite{wavelet_trans} or intermediate feature maps~\cite{freekd}, our method uniquely operates on the logit space. By applying DWT at the logit level, we eliminate the need to access or align high-dimensional intermediate features, resulting in a structurally simpler and more broadly applicable framework. Interpreting the logits vector as a semantic signal, we disentangle dominant class scores (low-frequency content) from subtle inter-class relationships (high-frequency detail), enabling targeted transfer of the teacher’s fine-grained decision-making patterns—information that is especially valuable for compact student models.

Importantly, FiGKD achieves strong performance not only on standard benchmarks such as CIFAR-100 and TinyImageNet, but also demonstrates effectiveness in fine-grained classification tasks including FGVR datasets, where distinguishing between visually similar classes is especially challenging.

\begin{table*}[htbp]
\caption{Top-1 Accuracy (\%) of various knowledge distillation methods on the CIFAR-100 validation set using homogeneous teacher-student model architectures. The highest accuracy is highlighted in \textbf{bold}, and the second-highest is \underline{underlined}.}
\centering
\resizebox{0.7\textwidth}{!}{%
    \begin{tabular}{@{}c|cccccccc@{}}
        \toprule
        \multirow{4}{*}{Types} & \multirow{2}{*}{Teacher}&\multicolumn{1}{c}{ResNet110} & \multicolumn{1}{c}{ResNet32x4}  & \multicolumn{1}{c}{WRN-40-2}  &   \multicolumn{1}{c}{WRN-40-2}    &\multicolumn{1}{c}{VGG13}  &  \multirow{4}{*}{Avg.} \\
        
         &&\multicolumn{1}{c}{74.31} & \multicolumn{1}{c}{79.42}    &\multicolumn{1}{c}{75.61}  &   \multicolumn{1}{c}{75.61}  &  \multicolumn{1}{c}{74.64}   \\   
         
        &\multirow{2}{*}{Student}& \multicolumn{1}{c}{ResNet32} & \multicolumn{1}{c}{ResNet8x4}  &  \multicolumn{1}{c}{WRN-16-2}  &   \multicolumn{1}{c}{WRN-40-1}  &  \multicolumn{1}{c}{VGG8}  \\ 
        
        &&  \multicolumn{1}{c}{71.14} & \multicolumn{1}{c}{72.50}  &  \multicolumn{1}{c}{73.26}  &   \multicolumn{1}{c}{71.98}  &  \multicolumn{1}{c}{70.36}  \\
        
        \midrule
        \multirow{11}{*}{Features} & FitNet~\cite{fitnet}                
        & 71.06 & 73.50 & 73.58 & 73.50 & 71.02 & 72.53 \\
        & PKT~\cite{pkt} & 72.61 & 73.64 & 74.54 & 73.64 & 72.88 & 73.46 \\
        & RKD~\cite{rkd} & 71.82 & 71.90 & 73.35 & 72.22 & 71.48 & 72.15 \\ 
        & CRD~\cite{crd} & 73.48 & 75.51 & 75.48 & \underline{75.51} & 73.94 & 74.78 \\ 
        & AT~\cite{at} & 72.31 & 73.44 & 74.08 & 73.44 & 71.43 & 72.94 \\ 
        & VID~\cite{vid} & 72.61 & 73.09 & 74.11 & 73.09 & 71.23 & 72.83 \\ 
        & OFD~\cite{ofd} & 73.23 & 74.95 & 75.24 & 74.95 & 73.95 & 74.46 \\ 
        & SP~\cite{sp} & 72.69 & 72.94 & 73.83 & 72.43 & 72.68 & 72.91 \\ 
        & DiffKD~\cite{diffkd} & - & 76.72 & - & 74.09 & - & 75.41 \\
        & ReviewKD~\cite{review} & 73.89 & 75.63 & 76.12 & 75.09 & 74.84 & 75.11 \\ 
        & CAT-KD~\cite{catkd} & 73.62 & 76.91 & 75.60 & 74.82 & 74.65 & 75.12 \\ 

        \midrule
        \multirow{10}{*}{Logits} 
        & KD~\cite{hinton}& 73.08 & 73.33 & 74.92 & 73.54 & 72.98 & 73.57 \\
        & DML~\cite{dml} & 72.03 & 72.12 & 73.58 & 72.68 & 71.79 & 72.44 \\
        & TAKD~\cite{takd} & 73.37 & 73.81 & 75.12 & 73.78 & 73.23 & 73.86 \\
        & DIST~\cite{dist} & - & 76.31 & - & 74.73 & - & 75.52 \\
        & DKD~\cite{dkd} & {74.11} & 76.32 & 76.24 & 74.81 & 74.68 & 75.23 \\
        & MLKD~\cite{mlkd} & {74.11} & {77.08} & \underline{76.63} & 75.35 & \underline{75.18} & \underline{75.67} \\

        & {RLD}~\cite{rld} & {74.02} & {76.64} & {-} & {74.88} & {74.93} & {75.12} \\

        & {TeKAP}~\cite{tekap} & {73.42} & {74.79} & {75.21} & {73.80} & {74.00} & {74.24} \\

        & {LDRLD}~\cite{ldrld} & {\underline{74.16}} & {{77.20}} & {76.35} & {74.98} & {75.06} & {75.55} \\

        & {EKD}~\cite{ekd} & {73.68} & {\underline{77.21}} & {76.15} & {74.43} & {74.71} & {75.24} \\

        \midrule
        \multirow{2}{*}{Ours} & \textbf{FiGKD} & \textbf{74.31} & \textbf{78.05} & \textbf{77.06} & \textbf{76.04} & \textbf{76.17} & \textbf{76.33} \\
        
        & $\Delta$ & \textbf{+0.15} & \textbf{+0.84} & \textbf{+0.43} & \textbf{+0.53} & \textbf{+0.99} & \textbf{+0.66} \\
        
        \bottomrule
    \end{tabular}
}
\label{tab:cifar_homogeneous}
\end{table*}

\begin{table*}[htbp]
\caption{Top-1 Accuracy (\%) of various knowledge distillation methods on the CIFAR-100 validation set using heterogeneous teacher-student model architectures. The highest accuracy is highlighted in \textbf{bold}, and the second-highest is \underline{underlined}.}
\centering
\resizebox{0.8\textwidth}{!}{%
    \begin{tabular}{@{}c|cccccccc@{}}
        \toprule
        \multirow{4}{*}{Types} & \multirow{2}{*}{Teacher}&  \multicolumn{1}{c}{WRN-40-2} & \multicolumn{1}{c}{ResNet50}  & \multicolumn{1}{c}{ResNet32x4}  &   \multicolumn{1}{c}{ResNet32x4}    &\multicolumn{1}{c}{VGG13} & \multirow{4}{*}{Avg.} \\ 
        
         && \multicolumn{1}{c}{75.61} & \multicolumn{1}{c}{79.34}    &\multicolumn{1}{c}{79.42}  &   \multicolumn{1}{c}{79.42}  &  \multicolumn{1}{c}{74.64}  \\
         
        &\multirow{2}{*}{Student}& \multicolumn{1}{c}{ShuffleNet-V1} & \multicolumn{1}{c}{MobileNet-V2}  &  \multicolumn{1}{c}{ShuffleNet-V1}  &   \multicolumn{1}{c}{ShuffleNet-V2}  &  \multicolumn{1}{c}{MobileNet-V2} \\ 
        
        &&  \multicolumn{1}{c}{70.50} & \multicolumn{1}{c}{64.60}  &  \multicolumn{1}{c}{70.50}  &   \multicolumn{1}{c}{71.82}  &  \multicolumn{1}{c}{64.60}  \\
        
        \midrule
        \multirow{11}{*}{Features} 
        & FitNet~\cite{fitnet}                
        & 73.73 & 63.16 & 73.59 & 73.54 & 64.14 & 69.63 \\
        & PKT~\cite{pkt} & 73.89 & 66.52 & 74.10 & 74.69 & 67.13 & 71.27 \\
        & RKD~\cite{rkd} & 72.21 & 64.43 & 72.28 & 73.21 &  64.52 & 69.33 \\
        & CRD~\cite{crd} & 76.05 & 69.11 & 75.11 & 75.65 & 69.73  & 73.13 \\        
        & AT~\cite{at} & 73.32 & 58.58 & 71.73 & 72.73 & 59.40 & 67.15 \\
        & VID~\cite{vid} & 73.61 & 67.57 & 73.38 & 73.40 &  65.56 & 70.70 \\
        & OFD~\cite{ofd} & 75.85 & 69.04 & 75.98 & 76.82 &  69.48  & 73.43 \\
        & SP~\cite{sp} & 74.52 & 68.08 & 73.48 & 74.56 & 66.30 & 71.39 \\
        & DiffKD~\cite{diffkd} &  - & 69.21 & 76.57 & 77.52 &  - & 74.43  \\ 
        & ReviewKD~\cite{review} & 77.14 & 69.89 & {77.45} & 77.78 &  70.37 & 74.53  \\
        & CAT-KD~\cite{catkd} & 77.35 & \underline{71.36} & \underline{78.26} & 78.41 & 69.13 & 74.90 \\

        \midrule
        \multirow{10}{*}{Logits} 
        & KD~\cite{hinton} & 74.83 & 67.35 & 74.07 & 74.45 & 67.37 & 71.61 \\
        & DML~\cite{dml} & 72.76 & 65.71 & 72.89 & 73.45 & 65.63 & 70.09 \\
        & TAKD~\cite{takd} & 75.34 & 68.02 & 74.53 & 74.82 & 67.91 & 72.12 \\
        & DIST~\cite{dist} & - & 68.66 & 76.34 & 77.35 & - & 74.12 \\ 
        & DKD~\cite{dkd} & 76.70 & 70.35 & 76.45 & 77.07 &  69.71 & 74.06 \\
        & MLKD~\cite{multi} & \underline{77.44} & 71.04 & 77.18 & \underline{78.44} & \underline{70.57} & \underline{74.93} \\ 

        & {RLD}~\cite{rld} & {-} & {70.76} & {-} & {77.56} & {69.97} & {72.76} \\
        
        & {TeKAP}~\cite{tekap} & {76.75} & {69.00} & {74.92} & {75.43} & {67.39} & {72.70} \\
        
        & {LDRLD}~\cite{ldrld} & {77.09} & {70.74} & {76.46} & {77.33} & {70.11} & {74.35} \\

        & {EKD}~\cite{ekd} & {-} & {70.38} & {-} & {77.65} & {69.94} & {72.66} & \\

        \midrule
        \multirow{2}{*}{Ours} & \textbf{FiGKD} & \textbf{78.84} & \textbf{72.08} & \textbf{78.70} & \textbf{79.64} & \textbf{71.81} & \textbf{76.21} \\
        & $\Delta$ & \textbf{+1.40} & \textbf{+0.72} & \textbf{+0.44} & \textbf{+1.20} & \textbf{+1.24} & \textbf{+1.28} \\
        
        \bottomrule
    \end{tabular}
}
\label{tab:cifar_heterogeneous}
\end{table*}

\section{Experiments}
\label{sec:experiment}

To validate the effectiveness of the proposed Fine-Grained Knowledge Distillation (FiGKD) framework, we conduct comprehensive experiments across a range of classification tasks. Our evaluation covers both coarse-grained benchmarks (CIFAR-100 and TinyImageNet) and fine-grained visual recognition (FGVR) datasets, including CUB-200, Stanford Dogs, MIT-67, and Stanford 40 Actions. These datasets exhibit varying levels of intra-class similarity and inter-class variation, making them well-suited for assessing FiGKD’s ability to transfer fine-grained semantic knowledge.

Unlike conventional logit-based KD methods that treat the teacher's output as a monolithic signal, FiGKD explicitly focuses on transferring high-frequency logit components that encode subtle decision boundaries and semantic detail. We hypothesize that our method will be particularly effective in FGVR scenarios, where successful classification often hinges on capturing minor but discriminative differences between visually similar classes.

We begin by describing the experimental setup, including dataset specifications, {implementation details}, and baseline KD methods used for comparison. We then present quantitative results across all datasets, demonstrating the effectiveness of FiGKD and its superiority over both logit-based and feature-based distillation techniques. Further analysis, including ablation study, visualization, {training overhead,} and teacher–student scale comparison, is provided in the subsequent Sec.~\ref{sec:discussion}.

\subsection{Datasets}
\label{subsec:datasets}

\subsubsection{Coarse-Grained Benchmarks} 
We employ CIFAR-100~\cite{cifar} and Tiny-ImageNet~\cite{imagenet} for standard classification evaluation. CIFAR-100 is a widely used image classification dataset consisting of 100 categories. Each image has a resolution of $32 \times 32$ pixels. The dataset contains 50,000 training images and 10,000 test images. Tiny-ImageNet is a subset of the full ImageNet dataset~\cite{imagenet}, comprising 200 object categories. All images are resized to $64 \times 64$ pixels. Each class includes 500 training images, 50 validation images, and 50 test images.

\subsubsection{Fine-Grained Visual Recognition (FGVR)}
To evaluate the effectiveness of our method on more challenging classification tasks that require fine-grained discrimination, we conduct experiments on the following FGVR benchmarks. For all FGVR datasets, we apply a \texttt{RandomResizedCrop}(224) transformation to standardize the input resolution while introducing spatial variation during training.
\begin{itemize}
    \item \textbf{Caltech-UCSD Birds (CUB200)~\cite{cub}}: Contains 200 bird species with 5,994 training and 5,794 test images. This dataset focuses on subtle visual differences among highly similar bird classes.  
    \item \textbf{MIT Indoor Scenes (MIT67)~\cite{mit67}}: Includes 67 indoor scene categories with 5,360 training and 1,340 test images. The task involves recognizing spatial layout and contextual cues across varied indoor environments.
    \item \textbf{Stanford 40 Actions (Stanford40)~\cite{stan40}}: Comprises 40 human action categories with a total of 9,532 images. This dataset emphasizes variations in human pose and object interaction.
    \item \textbf{Stanford Dogs~\cite{stan_dog}}: Consists of 120 dog breeds selected from ImageNet, with 12,000 training and 8,580 test images. The dataset is known for its high inter-class similarity, making it a particularly difficult fine-grained recognition task.
\end{itemize}

\subsection{Implementation Details}
\label{subsec:details}
We evaluate our method using widely adopted convolutional backbones, including VGG~\cite{vgg}, ResNet~\cite{resnet}, WideResNet (WRN)~\cite{wideresnet}, MobileNet~\cite{mobilenet}, and ShuffleNet~\cite{shufflenet, shufflenetv2}. Both homogeneous and heterogeneous teacher–student pairings are considered to assess the generality of our approach. 

All models are trained using stochastic gradient descent (SGD) with a momentum of 0.9 and a weight decay of $5 \times 10^{-4}$. The training is conducted for 240 epochs with a batch size of 64. The initial learning rate is set to 0.05, except for ShuffleNet-V1/V2 and MobileNet-V2, which use 0.01. The learning rate is decayed by a factor of 0.1 at epochs 150, 180, and 210. \textbf{Each experiment is repeated three times, and the average performance is reported.}

\subsection{Baselines}
\label{subsec:baseline}
We compare FiGKD against a comprehensive set of knowledge distillation methods, covering both logit-based and feature-based approaches:
\begin{itemize}
    \item \textbf{Logit-based baselines}: KD~\cite{hinton}, DML~\cite{dml}, TAKD~\cite{takd}, DIST~\cite{dist}, DKD~\cite{dkd}, MLKD~\cite{mlkd}, {RLD~\cite{rld}, TeKAP~\cite{tekap}, LDRLD~\cite{ldrld}, and EKD~\cite{ekd}}.  
    \item \textbf{Feature-based baselines}: FitNet~\cite{fitnet}, PKT~\cite{pkt}, RKD~\cite{rkd}, CRD~\cite{crd}, AT~\cite{at}, VID~\cite{vid}, OFD~\cite{ofd}, SP~\cite{sp}, DiffKD~\cite{diffkd}, ReviewKD~\cite{review}, and CAT-KD~\cite{catkd}.
\end{itemize}

\subsection{Main Results}
\label{subsec:classification}

\subsubsection{CIFAR-100}

\begin{table*}[htbp]
\caption{Top-1 Accuracy (\%) of various knowledge distillation methods on the TinyImageNet validation set using homogeneous teacher-student model architectures. The highest accuracy is highlighted in \textbf{bold}, and the second-highest is \underline{underlined}.}
\centering
\resizebox{0.70\linewidth}{!}{%
    \begin{tabular}{@{}c|cccccccc@{}}
        \toprule
        \multirow{4}{*}{Types} & \multirow{2}{*}{Teacher}&\multicolumn{1}{c}{ResNet110} & \multicolumn{1}{c}{ResNet32x4}  & \multicolumn{1}{c}{WRN-40-2}  &   \multicolumn{1}{c}{WRN-40-2}    &\multicolumn{1}{c}{VGG13}  &  \multirow{4}{*}{Avg.} \\
        
        &&\multicolumn{1}{c}{60.97} & \multicolumn{1}{c}{64.41} & \multicolumn{1}{c}{61.28}  &   \multicolumn{1}{c}{61.28}  & \multicolumn{1}{c}{62.59}   \\   
         
        &\multirow{2}{*}{Student}& \multicolumn{1}{c}{ResNet32} & \multicolumn{1}{c}{ResNet8x4}  &  \multicolumn{1}{c}{WRN-16-2}  &   \multicolumn{1}{c}{WRN-40-1}  &  \multicolumn{1}{c}{VGG8}  \\ 
        
        &&  \multicolumn{1}{c}{55.78} & \multicolumn{1}{c}{55.41}  &  \multicolumn{1}{c}{58.23}  &   \multicolumn{1}{c}{56.78}  &  \multicolumn{1}{c}{56.67}  \\
        
        \midrule
         \multirow{7}{*}{Features}
         & FitNet~\cite{fitnet} & 55.01 & 55.39 & 58.30 & 56.99 & 58.04 & 56.75 \\
         & PKT~\cite{pkt} & 57.81 & 56.25 & 59.04 & 58.02 & 58.73 & 57.97 \\
         & VID~\cite{vid} & 58.45 & 55.98 & 58.12 & 56.78 & 57.81 & 57.43 \\

         & AT~\cite{at} & 57.38 & 56.56 & 58.93 & 57.62 & 59.58 & 58.01 \\
         & RKD~\cite{ofd} & 57.23 & 54.24 & 57.64 & 56.07 & 58.15 & 56.67 \\
         & ReviewKD~\cite{review} & 59.24 & 59.01 & 60.55 & \underline{60.27} & 62.03 & 59.42 \\
         & CAT-KD~\cite{catkd} & 58.58 & 59.11 & 59.60 & 59.67 & 62.66 & \underline{59.92} \\

         \midrule
        \multirow{3}{*}{Logits} 
        & KD~\cite{hinton} & 57.64 & 55.41 & 58.23 & 58.25 & 56.67 & 57.24 \\
        & DKD~\cite{dkd} & \underline{59.31} & 55.67 & 58.65 & 58.94 & 61.48 & 58.81 \\
        & MLKD~\cite{mlkd} & 58.30 & \underline{60.67} & \underline{60.80} & 59.91 & \underline{63.76} & 59.70 \\

         \midrule
        \multirow{2}{*}{Ours} & \textbf{FiGKD} & \textbf{59.73} & \textbf{60.92} & \textbf{61.91} & \textbf{60.69} & \textbf{64.52} & \textbf{60.51} \\
        & $\Delta$ & \textbf{+0.42} & \textbf{+0.25} & \textbf{+1.11} & \textbf{+0.42} & \textbf{+0.76} & \textbf{+0.58} \\
        
        \bottomrule
    \end{tabular}
}
\label{tab:tiny_homogeneous}
\end{table*}

We first evaluate the effectiveness of FiGKD on the CIFAR-100 benchmark under both homogeneous and heterogeneous teacher–student configurations. The results are summarized in Tables~\ref{tab:cifar_homogeneous} and~\ref{tab:cifar_heterogeneous}, with baseline performances reproduced from original papers or public implementations where available.

Table~\ref{tab:cifar_homogeneous} presents the results in the homogeneous setting, where the teacher and student share the same backbone family (e.g., ResNet-to-ResNet, WRN-to-WRN). While feature-based KD methods have traditionally achieved strong performance by leveraging rich intermediate representations, recent logit-based approaches have significantly narrowed this gap. Even with these advances, FiGKD consistently achieves the highest accuracy across all teacher–student pairs, outperforming even the strongest feature-based baselines, including DiffKD, ReviewKD, and CAT-KD. In particular, FiGKD yields an average top-1 accuracy improvement of +0.66\%, with gains of +0.97\% over MLKD for ResNet32x4 $\rightarrow$ ResNet8x4 and +0.99\% for VGG13 $\rightarrow$ VGG8. These results demonstrate the effectiveness of selectively distilling high-frequency logit components that capture fine-grained class relationships.

Table~\ref{tab:cifar_heterogeneous} presents the results for the heterogeneous setting, where the student model architecture differs from that of the teacher (e.g., ResNet-to-MobileNet, WRN-to-ShuffleNet), reflecting realistic deployment scenarios under strict resource constraints. Despite the increased difficulty, FiGKD consistently outperforms all compared methods. It surpasses the average performance of strong feature-based baselines such as ReviewKD and CAT-KD, as well as advanced logit-based methods like MLKD. For instance, FiGKD achieves improvements of +1.24\% over MLKD in the VGG13 $\rightarrow$ MobileNetV2 setting and +1.40\% in WRN-40-2 $\rightarrow$ ShuffleNetV1, resulting in an overall average gain of +1.28\% across all heterogeneous tasks.

Importantly, although FiGKD operates solely on the teacher’s final output logits, it consistently outperforms methods that require access to internal feature maps. This confirms that frequency-aware logit decomposition can isolate and transfer the most informative teacher knowledge, without the architectural coupling or memory overhead typical of feature-based approaches.

\subsubsection{TinyImageNet}

\begin{table*}[htbp]
\caption{Top-1 Accuracy (\%) of various knowledge distillation methods on the TinyImageNet validation set using heterogeneous teacher-student model architectures. The highest accuracy is highlighted in \textbf{bold}, and the second-highest is \underline{underlined}.}
\centering
\resizebox{0.80\linewidth}{!}{%
    \begin{tabular}{@{}ccccccccc@{}}
        \toprule
        \multirow{4}{*}{Types} & \multirow{2}{*}{Teacher}&\multicolumn{1}{c}{WRN-40-2} & \multicolumn{1}{c}{ResNet50}  & \multicolumn{1}{c}{ResNet32x4} & \multicolumn{1}{c}{ResNet32x4} &\multicolumn{1}{c}{VGG13} & \multirow{4}{*}{Avg.} \\
        
        &&\multicolumn{1}{c}{61.28} & \multicolumn{1}{c}{68.20} & \multicolumn{1}{c}{64.41}  &   \multicolumn{1}{c}{64.41}  & \multicolumn{1}{c}{62.59}   \\   
         
        & \multirow{2}{*}{Student}& \multicolumn{1}{c}{ShuffleNet-V1} & \multicolumn{1}{c}{MobileNet-V2}  &  \multicolumn{1}{c}{ShuffleNet-V1} & \multicolumn{1}{c}{ShuffleNet-V2} & \multicolumn{1}{c}{MobileNet-V2}  \\ 
        
        &&  \multicolumn{1}{c}{58.20} & \multicolumn{1}{c}{58.20}  &  \multicolumn{1}{c}{58.20}  &   \multicolumn{1}{c}{62.07}  &  \multicolumn{1}{c}{58.20}  \\
        
        \midrule
        \multirow{7}{*}{Features}
        & FitNet~\cite{fitnet} & 57.33 & 57.54 & 57.74 & 64.70 & 57.90 & 59.04 \\
        & PKT~\cite{pkt} & 59.65 & 59.57 & 59.61 & 65.53 & 59.70 & 60.81 \\
        & VID~\cite{vid} & 58.26 & 58.27 & 58.07 & 64.50 & 58.53 & 59.53 \\

        & AT~\cite{at} & 57.89 & 51.27 & 56.57 & 65.61 & 57.78 & 57.82 \\
        & RKD~\cite{ofd} & 59.20 & 58.33 & 58.21 & 64.33 & 58.25 & 59.66 \\
        & ReviewKD~\cite{review} & \underline{63.29} & 59.35 & \underline{64.47} & 67.89 & 62.06 & 63.41 \\
        & CAT-KD~\cite{catkd} & 61.43 & 58.57 & - & 67.69 & 57.42 & 61.28 \\

        \midrule
        \multirow{3}{*}{Logits} 
        & KD~\cite{hinton} & 61.10 & 58.79 & 61.73 & 62.07 & 58.20 & 60.38 \\
        & DKD~\cite{dkd} & 61.64 & 61.42 & 62.58 & 66.34 & 59.28 & 62.25 \\
        & MLKD~\cite{mlkd} & 62.85 & \underline{62.42} & 63.80 & \underline{68.63} & \underline{63.16} & \underline{64.17} \\

        \midrule
        \multirow{2}{*}{Ours} 
        & \textbf{FiGKD} & \textbf{64.95} & \textbf{63.68} & \textbf{65.84} & \textbf{70.10} & \textbf{63.99} & \textbf{65.71} \\
        & $\Delta$ & \textbf{+1.66} & \textbf{+1.26} & \textbf{+1.37} & \textbf{+1.47} & \textbf{+0.83} & \textbf{+1.54} \\
        
        \bottomrule
    \end{tabular}
}
\label{tab:tiny_heterogeneous}
\end{table*}

We further evaluate the performance of FiGKD on the TinyImageNet dataset, which presents a more complex classification challenge due to its larger number of categories and greater visual variability. Experimental results are shown in Tables~\ref{tab:tiny_homogeneous} and~\ref{tab:tiny_heterogeneous}, covering both homogeneous and heterogeneous teacher–student configurations. All baseline results are reproduced from original papers or publicly available implementations, where applicable.

In the homogeneous setting (Table~\ref{tab:tiny_homogeneous}), where the teacher and student share the same backbone family, FiGKD achieves the highest accuracy across all model pairs. Compared to the strongest logit-based baseline (MLKD), FiGKD yields an average improvement of +0.58\%, including gains of +1.11\% for WRN-40-2 $\rightarrow$ WRN-16-2 and +0.76\% for VGG13 $\rightarrow$ VGG8. FiGKD also outperforms advanced feature-based methods such as ReviewKD and CAT-KD. These results confirm the advantage of distilling high-frequency logit components that encode fine-grained semantic details—particularly beneficial for visually complex tasks.

In the heterogeneous setting (Table~\ref{tab:tiny_heterogeneous}), where the student model architecture differs from that of the teacher (e.g., WRN-40-2 $\rightarrow$ ShuffleNet-V1, ResNet50 $\rightarrow$ MobileNetV2), FiGKD demonstrates robust performance across heterogeneous architectures. Despite the increased difficulty introduced by architectural mismatch, it achieves the highest top-1 accuracy across all configurations, with an average gain of +1.54\% over MLKD. For instance, FiGKD improves performance by +1.66\% in WRN-40-2 $\rightarrow$ ShuffleNet-V1 and +1.47\% in ResNet32x4 $\rightarrow$ ShuffleNet-V2. These results highlight the robustness of FiGKD’s frequency-based distillation strategy in transferring fine-grained semantic knowledge across diverse architectural settings.

\begin{table*}[t]
\caption{Top-1 Accuracy (\%) of various knowledge distillation methods on the FGVR validation set using homogeneous and heterogeneous teacher-student model architectures. The highest accuracy is highlighted in \textbf{bold}, and the second-highest is \underline{underlined}.}
\centering
\resizebox{0.85\linewidth}{!}{%
    \begin{tabular}{@{}ccccccccc@{}}
        \toprule
        Dataset & \multicolumn{2}{c}{CUB200} &  \multicolumn{2}{c}{MIT67} &  \multicolumn{2}{c}{Stanford40} & \multicolumn{2}{c}{Dogs}\\ 
        \midrule
        
        \multirow{2}{*}{Teacher}&  \multicolumn{1}{c}{ResNet34} & \multicolumn{1}{c}{MobileNetV1}  & \multicolumn{1}{c}{ResNet34}  &   \multicolumn{1}{c}{MobileNetV1}    &\multicolumn{1}{c}{ResNet34}  &   \multicolumn{1}{c}{MobileNetV1}    &\multicolumn{1}{c}{ResNet34}  &   \multicolumn{1}{c}{MobileNetV1}    \\ 
         & \multicolumn{1}{c}{61.43} & \multicolumn{1}{c}{67.02}    &\multicolumn{1}{c}{59.55}  &   \multicolumn{1}{c}{61.64}  &  \multicolumn{1}{c}{49.06}  &   \multicolumn{1}{c}{56.06}  &  \multicolumn{1}{c}{69.28}  &   \multicolumn{1}{c}{69.83} \\
        
        \multirow{2}{*}{Student}& \multicolumn{1}{c}{ResNet18} & \multicolumn{1}{c}{ResNet18}  &  \multicolumn{1}{c}{ResNet18}  &   \multicolumn{1}{c}{ResNet18}  &  \multicolumn{1}{c}{ResNet18}  &   \multicolumn{1}{c}{ResNet18}  &  \multicolumn{1}{c}{ResNet18}  &   \multicolumn{1}{c}{ResNet18}    \\ 
        &  \multicolumn{1}{c}{58.14} & \multicolumn{1}{c}{58.14}  &  \multicolumn{1}{c}{57.49}  &   \multicolumn{1}{c}{57.49}  &  \multicolumn{1}{c}{45.94}  &   \multicolumn{1}{c}{45.94}  &  \multicolumn{1}{c}{66.97}  &   \multicolumn{1}{c}{66.97} \\
        \midrule
        
        FitNet~\cite{fitnet}                      
        & 59.60 & 56.00 & 58.28 & 57.07 &  46.89 &44.04 &  67.06 & 66.25\\
        PKT~\cite{pkt}                
        & 60.98 & 64.71 & 60.80 & 64.73 & 50.25 & 54.18 & 69.62 & 71.72 \\
        RKD~\cite{rkd}                
        &  54.80 & 58.80 & 57.63 & 62.14 &  46.68 & 51.12 &  67.23 & 70.49 \\
        CRD~\cite{crd}                     
        &  60.29 & 64.53 & 59.70 & 63.92 &  49.77 & 54.26 &  68.67 & 70.98 \\
        AT~\cite{at}                     
        & 63.48 & 66.66 & 60.60 & 61.22 &  51.08 & 55.51 & 70.05 & 71.34 \\
        ReviewKD~\cite{review}                      
        &  62.13 & 63.09 & 59.68 & 60.76 &  49.95 & 51.77 &  68.96 & 69.22 \\
        \midrule
        KD~\cite{hinton}                    
        &  60.92 & 64.74 & 58.78 & 61.87 & 49.42 & 54.07 &  68.28 & 71.82 \\
        DKD~\cite{dkd}               
        &  62.17 & 66.45 & 60.00 & 64.35 & 49.84 & 55.80  & 69.04 & 72.53 \\
        MLKD~\cite{mlkd}               
        &  \underline{64.77} & \underline{69.62} & \underline{61.89} & \underline{65.35} & \underline{52.86} & \underline{59.06} & \underline{71.81} & \underline{73.51} \\
        \midrule
 
        \textbf{FiGKD}   
        &  \textbf{67.93} & \textbf{71.97} & \textbf{65.67} & \textbf{68.43} & \textbf{53.69} & \textbf{60.21} & \textbf{72.79} & \textbf{74.38}\\ 
        $\Delta$ 
        &  \textbf{+3.16} & \textbf{+2.35} & \textbf{+3.78} & \textbf{+3.08} &  \textbf{+0.83} & \textbf{+1.15} &  \textbf{+0.98} & \textbf{+0.87}\\
        \bottomrule
    \end{tabular}
}
\label{Table:FGVR}
\end{table*}

\subsubsection{{FGVR}}

Lastly, we evaluate FiGKD on fine-grained visual recognition (FGVR) benchmarks, where the objective is to distinguish between highly similar classes based on subtle visual cues. These tasks are especially challenging for compact student models, which often lack the capacity to replicate the nuanced decision boundaries learned by larger teacher networks.

Table~\ref{Table:FGVR} presents the results on four widely used FGVR datasets: CUB200, MIT67, Stanford40, and Stanford Dogs. Across all datasets and both homogeneous (ResNet34 $\rightarrow$ ResNet18) and heterogeneous (MobileNetV1 $\rightarrow$ ResNet18) settings, FiGKD consistently achieves the highest accuracy, outperforming both logit-based and feature-based baselines.

\begin{table*}[]
\caption{{Top-1 Accuracy (\%) of various methods on CUB200 with different teacher-student architectures. The best accuracy is highlighted in \textbf{bold}, and the second-best is \underline{underlined}.}}
\centering
\resizebox{0.9\linewidth}{!}{%
    \begin{tabular}{c|cccc|cccc}
    
        \toprule
        \multirow{2}{*}{Teacher} &
        \multicolumn{1}{c}{ResNet32x4} & \multicolumn{1}{c}{WRN-40-2}  & \multicolumn{1}{c}{WRN-40-2}  &   \multicolumn{1}{c|}{VGG13} &
        \multicolumn{1}{c}{WRN-40-2} &  \multicolumn{1}{c}{ResNet32x4} &
        \multicolumn{1}{c}{ResNet32x4} &
        \multicolumn{1}{c}{VGG13}
        \\

        &
        \multicolumn{1}{c}{55.13} & 
        \multicolumn{1}{c}{52.38} &   \multicolumn{1}{c}{52.38} & 
        \multicolumn{1}{c|}{48.12} & 
        \multicolumn{1}{c}{52.38} & 
        \multicolumn{1}{c}{55.13} &   \multicolumn{1}{c}{55.13} & 
        \multicolumn{1}{c}{48.12}   
        \\   
         
        \multirow{2}{*}{Student} &
        \multicolumn{1}{c}{ResNet8x4} & \multicolumn{1}{c}{WRN-16-2}  & \multicolumn{1}{c}{WRN-40-1}  &   \multicolumn{1}{c|}{VGG8} &
        \multicolumn{1}{c}{ShuffleNet-V1} &  \multicolumn{1}{c}{ShuffleNet-V1} &
        \multicolumn{1}{c}{ShuffleNet-V2} &
        \multicolumn{1}{c}{MobileNet-V2}
        \\

        &
        \multicolumn{1}{c}{42.09} & 
        \multicolumn{1}{c}{50.11} &
        \multicolumn{1}{c}{47.93} & 
        \multicolumn{1}{c|}{44.19} & 
        \multicolumn{1}{c}{21.23} & 
        \multicolumn{1}{c}{21.23} &
        \multicolumn{1}{c}{38.37} & 
        \multicolumn{1}{c}{23.43}   
        \\   
        
        \midrule
        KD~\cite{hinton} & 53.94 & 55.95 & 53.96 & 50.99 & 36.43 & 34.92 & 49.85 & 38.29 \\ 
        DKD~\cite{dkd} & 55.61 & 55.36 & 54.09 & 50.60 & 40.51 & 40.15 & 53.24 & 44.11 \\
        MLKD~\cite{mlkd} & \underline{57.18} & \underline{56.62} & \underline{56.10} & \underline{51.75} & \underline{45.21} & \underline{42.78} & \underline{55.18} & \underline{45.13} \\

        \midrule
        \textbf{FiGKD} & \textbf{58.84} & \textbf{57.49} & \textbf{56.35} & \textbf{52.13} & \textbf{46.20} & \textbf{43.36} & \textbf{56.81} & \textbf{46.49} \\

        $\Delta$ & \textbf{+1.66} & \textbf{+0.87} & \textbf{+0.25} & \textbf{+0.38} & \textbf{+0.99} & \textbf{+0.58} & \textbf{+1.63} & \textbf{+1.36} \\
        
        \bottomrule
    \end{tabular}
}
\label{tab:cub200}
\end{table*}

\begin{table*}[]
\caption{{Top-1 Accuracy (\%) of various methods on MIT67 with different teacher-student architectures. The best accuracy is highlighted in \textbf{bold}, and the second-best is \underline{underlined}.}}
\centering
\resizebox{0.9\linewidth}{!}{%
    \begin{tabular}{c|cccc|cccc}
    
        \toprule
        \multirow{2}{*}{Teacher} &
        \multicolumn{1}{c}{ResNet32x4} & \multicolumn{1}{c}{WRN-40-2}  & \multicolumn{1}{c}{WRN-40-2}  &   \multicolumn{1}{c|}{VGG13} &
        \multicolumn{1}{c}{WRN-40-2} &  \multicolumn{1}{c}{ResNet32x4} &
        \multicolumn{1}{c}{ResNet32x4} &
        \multicolumn{1}{c}{VGG13}
        \\

        &
        \multicolumn{1}{c}{55.30} & 
        \multicolumn{1}{c}{53.51} &   \multicolumn{1}{c}{53.51} & 
        \multicolumn{1}{c|}{50.15} & 
        \multicolumn{1}{c}{53.51} & 
        \multicolumn{1}{c}{55.30} &   \multicolumn{1}{c}{55.30} & 
        \multicolumn{1}{c}{50.15}   
        \\   
         
        \multirow{2}{*}{Student} &
        \multicolumn{1}{c}{ResNet8x4} & \multicolumn{1}{c}{WRN-16-2}  & \multicolumn{1}{c}{WRN-40-1}  &   \multicolumn{1}{c|}{VGG8} &
        \multicolumn{1}{c}{ShuffleNet-V1} &  \multicolumn{1}{c}{ShuffleNet-V1} &
        \multicolumn{1}{c}{ShuffleNet-V2} &
        \multicolumn{1}{c}{MobileNet-V2}
        \\

        &
        \multicolumn{1}{c}{53.08} & 
        \multicolumn{1}{c}{53.01} &
        \multicolumn{1}{c}{49.50} & 
        \multicolumn{1}{c|}{47.54} & 
        \multicolumn{1}{c}{30.84} & 
        \multicolumn{1}{c}{30.84} &
        \multicolumn{1}{c}{40.99} & 
        \multicolumn{1}{c}{33.13}   
        \\   
        
        \midrule
        KD~\cite{hinton} & 56.97 & \underline{56.57} & 52.79 & 52.56 & 44.53 & 42.24 & 53.38 & 41.37 \\ 
        DKD~\cite{dkd} & 57.98 & 55.75 & 53.13 & 53.21 & \underline{48.31} & \underline{48.80} & 56.27 & 44.07 \\
        MLKD~\cite{mlkd} & \underline{59.45} & 55.32 & \underline{53.98} & \underline{54.03} & 44.70 & 45.72 & \underline{57.95} & \underline{46.91} \\

        \midrule
        \textbf{FiGKD} & \textbf{60.30} & \textbf{57.44} & \textbf{56.34} & \textbf{54.58} & \textbf{50.35} & \textbf{50.25} & \textbf{58.29} & \textbf{48.25} \\

        $\Delta$ & \textbf{+0.85} & \textbf{+0.87} & \textbf{+2.36} & \textbf{+0.55} & \textbf{+2.04} & \textbf{+1.45} & \textbf{+0.34} & \textbf{+1.34} \\
        
        \bottomrule
    \end{tabular}
}
\label{tab:mit}
\end{table*}

\begin{table*}[]
\caption{{Top-1 Accuracy (\%) of various methods on Stanford40 with different teacher-student architectures. The best accuracy is highlighted in \textbf{bold}, and the second-best is \underline{underlined}.}}
\centering
\resizebox{0.9\linewidth}{!}{%
    \begin{tabular}{c|cccc|cccc}
    
        \toprule
        \multirow{2}{*}{Teacher} &
        \multicolumn{1}{c}{ResNet32x4} & \multicolumn{1}{c}{WRN-40-2}  & \multicolumn{1}{c}{WRN-40-2}  &   \multicolumn{1}{c|}{VGG13} &
        \multicolumn{1}{c}{WRN-40-2} &  \multicolumn{1}{c}{ResNet32x4} &
        \multicolumn{1}{c}{ResNet32x4} &
        \multicolumn{1}{c}{VGG13}
        \\

        &
        \multicolumn{1}{c}{46.20} & 
        \multicolumn{1}{c}{44.04} &   
        \multicolumn{1}{c}{44.04} & 
        \multicolumn{1}{c|}{42.82} & 
        \multicolumn{1}{c}{44.04} & 
        \multicolumn{1}{c}{46.20} &   
        \multicolumn{1}{c}{46.20} & 
        \multicolumn{1}{c}{42.82}   
        \\   
         
        \multirow{2}{*}{Student} &
        \multicolumn{1}{c}{ResNet8x4} & \multicolumn{1}{c}{WRN-16-2}  & \multicolumn{1}{c}{WRN-40-1}  &   \multicolumn{1}{c|}{VGG8} &
        \multicolumn{1}{c}{ShuffleNet-V1} &  \multicolumn{1}{c}{ShuffleNet-V1} &
        \multicolumn{1}{c}{ShuffleNet-V2} &
        \multicolumn{1}{c}{MobileNet-V2}
        \\

        &
        \multicolumn{1}{c}{37.23} & 
        \multicolumn{1}{c}{40.62} &
        \multicolumn{1}{c}{40.15} & 
        \multicolumn{1}{c|}{37.73} & 
        \multicolumn{1}{c}{24.87} & 
        \multicolumn{1}{c}{24.87} &
        \multicolumn{1}{c}{34.98} & 
        \multicolumn{1}{c}{26.01}   
        \\   
        
        \midrule
        KD~\cite{hinton} & 41.53 & 46.46 & 45.25 & 43.10 & 35.22 & 32.58 & 42.64 & 33.75 \\ 
        DKD~\cite{dkd} & 44.94 & 47.41 & 46.38 & 44.56 & \underline{39.92} & \underline{37.80} & 46.99 & 37.87 \\
        MLKD~\cite{mlkd} & \underline{45.95} & \underline{47.94} & \underline{46.78} & \underline{45.48} & 39.16 & 36.38 & \underline{47.92} & \underline{40.35} \\

        \midrule
        \textbf{FiGKD} & \textbf{47.26} & \textbf{48.28} & \textbf{47.27} & \textbf{46.51} & \textbf{40.99} & \textbf{38.93} & \textbf{48.99} & \textbf{40.65} \\

        $\Delta$ & \textbf{+1.31} & \textbf{+0.34} & \textbf{+0.49} & \textbf{+1.03} & \textbf{+1.07} & \textbf{+1.13} & \textbf{+1.07} & \textbf{+0.30} \\
        
        \bottomrule
    \end{tabular}
}
\label{tab:stanford}
\end{table*}

\begin{table*}[]
\caption{{Top-1 Accuracy (\%) of various methods on Dogs with different teacher-student architectures. The best accuracy is highlighted in \textbf{bold}, and the second-best is \underline{underlined}.}}
\centering
\resizebox{0.9\linewidth}{!}{%
    \begin{tabular}{c|cccc|cccc}
    
        \toprule
        \multirow{2}{*}{Teacher} &
        \multicolumn{1}{c}{ResNet32x4} & \multicolumn{1}{c}{WRN-40-2}  & \multicolumn{1}{c}{WRN-40-2}  &   \multicolumn{1}{c|}{VGG13} &
        \multicolumn{1}{c}{WRN-40-2} &  \multicolumn{1}{c}{ResNet32x4} &
        \multicolumn{1}{c}{ResNet32x4} &
        \multicolumn{1}{c}{VGG13}
        \\

        &
        \multicolumn{1}{c}{67.79} & 
        \multicolumn{1}{c}{65.64} &   
        \multicolumn{1}{c}{65.64} & 
        \multicolumn{1}{c|}{60.55} & 
        \multicolumn{1}{c}{65.64} & 
        \multicolumn{1}{c}{67.79} &   
        \multicolumn{1}{c}{67.79} & 
        \multicolumn{1}{c}{60.55}   
        \\   
         
        \multirow{2}{*}{Student} &
        \multicolumn{1}{c}{ResNet8x4} & \multicolumn{1}{c}{WRN-16-2}  & \multicolumn{1}{c}{WRN-40-1}  &   \multicolumn{1}{c|}{VGG8} &
        \multicolumn{1}{c}{ShuffleNet-V1} &  \multicolumn{1}{c}{ShuffleNet-V1} &
        \multicolumn{1}{c}{ShuffleNet-V2} &
        \multicolumn{1}{c}{MobileNet-V2}
        \\

        &
        \multicolumn{1}{c}{58.62} & 
        \multicolumn{1}{c}{62.51} &
        \multicolumn{1}{c}{62.64} & 
        \multicolumn{1}{c|}{53.34} & 
        \multicolumn{1}{c}{39.40} & 
        \multicolumn{1}{c}{39.40} &
        \multicolumn{1}{c}{57.26} & 
        \multicolumn{1}{c}{46.14}   
        \\   
        
        \midrule
        KD~\cite{hinton} & 63.43 & 67.07 & 67.03 & 60.33 & 56.70 & 53.75 & 66.37 & 57.26 \\ 
        DKD~\cite{dkd} & 66.23 & 66.96 & 66.51 & 61.98 & 59.66 & \underline{58.62} & 67.97 & 59.57 \\
        MLKD~\cite{mlkd} & \underline{67.07} & \underline{67.50} & \underline{67.61} & \underline{62.58} & \underline{60.90} & 57.57 & \underline{69.54} & \underline{60.56} \\

        \midrule
        \textbf{FiGKD} & \textbf{67.27} & \textbf{68.53} & \textbf{68.48} & \textbf{63.41} & \textbf{61.93} & \textbf{61.32} & \textbf{70.14} & \textbf{61.00} \\

        $\Delta$ & \textbf{+0.20} & \textbf{+1.03} & \textbf{+0.87} & \textbf{+0.83} & \textbf{+1.03} & \textbf{+2.70} & \textbf{+0.60} & \textbf{+0.44} \\
        
        \bottomrule
    \end{tabular}
}
\label{tab:dogs}
\end{table*}

Notably, FiGKD achieves substantial gains on CUB200 and MIT67—two datasets that demand fine-grained semantic discrimination—with improvements of +3.16\% and +3.78\% over the strongest baseline, respectively. Even under heterogeneous configurations, FiGKD shows notable gains of +2.35\% on CUB200 and +3.08\% on MIT67, demonstrating its robustness across different architectures.

To further verify the robustness of FiGKD regardless of input resolution and model architecture, we conducted additional experiments on FGVR datasets by resizing images to $64 \times 64$ and employing the same teacher–student pairs used in CIFAR-100 and TinyImageNet. The results are summarized in Tables~\ref{tab:cub200}, \ref{tab:mit}, \ref{tab:stanford}, and \ref{tab:dogs}.
As observed, FiGKD consistently outperforms other distillation methods across all  configurations. Notably, on the challenging MIT67 and Stanford Dogs datasets, FiGKD achieves substantial gains of up to +2.36\% and +2.70\% over the strongest baselines, respectively. These results confirm that the effectiveness of FiGKD stems from its frequency-aware distillation mechanism rather than specific high-resolution architectures, demonstrating its strong generalization capability even under constrained input settings.

Unlike prior methods that distill the full logit vector or rely on intermediate feature representations, FiGKD selectively transfers high-frequency logit components that encode fine-grained semantic cues. This targeted strategy proves especially effective for FGVR tasks, where subtle distinctions between classes are crucial. By doing so, FiGKD enables compact student models to retain fine-grained discriminative capacity, making it well-suited for real-world applications that demand both efficiency and accuracy.

\begin{table*}[htbp]
\caption{Ablation study on frequency components.}
\centering
\resizebox{0.75\linewidth}{!}{%
\begin{tabular}{c|cc|ccc}
\toprule
Datasets & \multicolumn{2}{c|}{Frequency} & \multicolumn{3}{c}{Accuracy (\%)} \\ \midrule
\multirow{5.5}{*}{CIFAF100} & Low & High & WRN-40-2 $\rightarrow$ WRN-16-2 & VGG13 $\rightarrow$ VGG8 & VGG13 $\rightarrow$ MobileNet-V2 \\ \cmidrule{2-6}

& \xmark & \xmark & 76.63 & 75.18 & 70.57 \\
& \cmark & \xmark & 76.44 & 75.23 & 70.65  \\
& \xmark & \cmark & \textbf{77.06} & \textbf{76.17} & \textbf{71.81}  \\
& \cmark & \cmark & 76.56 & 75.83 & 71.18  \\ \midrule

\multirow{5.5}{*}{TinyImageNet} & Low & High & ResNet32x4 $\rightarrow$ ResNet8x4 & VGG13 $\rightarrow$ VGG8 & ResNet32x4 $\rightarrow$ ShuffleNet-V2 \\ \cmidrule{2-6}

& \xmark & \xmark & 60.67 & 63.76 & 68.63  \\
& \cmark & \xmark & 60.45 & 63.99 & 69.39 \\
& \xmark & \cmark & \textbf{60.92} & \textbf{64.52} & \textbf{70.10}  \\
& \cmark & \cmark & 60.71 & 64.19 & 69.59 \\ \bottomrule
\end{tabular}
}
\label{Table:frequency}
\end{table*}

\section{Discussion}
\label{sec:discussion}
In Sec.~\ref{sec:experiment}, we demonstrated that FiGKD consistently outperforms both logit-based and feature-based baselines across a wide range of benchmarks and architectural configurations. To gain deeper insight into the underlying mechanisms driving this strong performance, we conduct a series of additional analyses, including ablation studies, visualization experiments, and model complexity evaluations.

\subsection{Ablation Study}
\label{subsec:ablation_study}

\subsubsection{Frequency Effect}

To investigate the effect of different frequency components, we conduct an ablation study that isolates the contributions of low- and high-frequency information in the wavelet-decomposed logit space. Table~\ref{Table:frequency} demonstrates the results for several representative teacher–student pairs on CIFAR-100 and TinyImageNet. Specifically, we compare four variants: (1) distilling neither frequency (Low~\xmark, High~\xmark), which means the baseline model, (2) distilling only the low-frequency (Low~\cmark, High~\xmark), (3) distilling only the high-frequency (Low~\xmark, High~\cmark), and (4) distilling both frequency (Low~\cmark, High~\cmark).

The results clearly show that distilling only the high-frequency logit components consistently yields the best performance across all settings. For example, on CIFAR-100, transferring only the high-frequency information achieves 77.06\% accuracy for WRN-40-2 $\rightarrow$ WRN-16-2, 76.17\% for VGG13 $\rightarrow$ VGG8 and 71.81\% for VGG13 $\rightarrow$ MobileNet-V2. Similarly, on TinyImageNet, the high-frequency-only variant reaches 60.92\%, 64.52\%, and 70.10\% for ResNet32x4 $\rightarrow$ ResNet8x4, VGG13 $\rightarrow$ VGG8, and ResNet32x4 $\rightarrow$ ShuffleNet-V2, respectively.

In contrast, distilling only the low-frequency component provides marginal improvements or even degrades performance compared to the baseline. While combining both frequencies yields better results than low-frequency distillation alone, it still underperforms compared to using only the high-frequency component.

These results highlight the importance of selectively transferring the teacher’s high-frequency logit information, which captures subtle, class-discriminative patterns essential for fine-grained recognition. By focusing on these high-frequency components, FiGKD enables the student model to better emulate the nuanced decision-making of the teacher while avoiding potential redundancy or noise present in the low-frequency content.

\subsubsection{Hyperparameter}

\begin{figure}[tbp]
    \centering
    \footnotesize
    \begin{tikzpicture}
    \begin{axis}[
        xlabel={Loss Weight},
        ylabel={Accuracy (\%)},
        ymin=77.0, ymax=78.2,
        xtick={1,2,4,6,8},
        ytick={77.2,77.4,77.6,77.8,78.0},
        grid=major,
        width=0.85\columnwidth,
        height=5cm,
        legend style={
            at={(0.5,1.12)},
            anchor=south,
            cells={align=left},
            draw=none,
        },
        legend cell align={left}
    ]

    \addplot[
        mark=o,
        thick,
        blue
    ] coordinates {
        (1,77.22)
        (2,77.84)
        (4,77.79)
        (6,77.72)
        (8,77.73)
    };
    \addlegendentry{Exp. 1: $\alpha=1, \beta=\{1,2,4,6,8\}$}
    \node at (axis cs:2,77.84) [anchor=south west, font=\footnotesize, fill=white] {77.84};

    \addplot[
        mark=square*,
        thick,
        orange
    ] coordinates {
        (1,77.84)
        (2,78.05)
        (4,78.02)
        (6,77.91)
        (8,77.73)
    };
    \addlegendentry{Exp. 2: $\beta=2, \alpha=\{1,2,4,6,8\}$}
    \node at (axis cs:2,78.05) [anchor=south, font=\footnotesize, fill=white] {78.05};

    \end{axis}

    
    \end{tikzpicture}

    \begin{tikzpicture}
    \begin{axis}[
        xlabel={Loss Weight},
        ylabel={Accuracy (\%)},
        ymin=59.4, ymax=61.4,
        xtick={1,2,4,6,8},
        ytick={59.8, 60.2, 60.6, 61.0},
        grid=major,
        width=0.85\columnwidth,
        height=5cm,
        legend style={
            at={(0.5,1.12)},
            anchor=south,
            cells={align=left},
            draw=none,
        },
        legend cell align={left}
    ]

    \addplot[
        mark=o,
        thick,
        blue
    ] coordinates {
        (1,59.74)
        (2,59.86)
        (4,59.80)
        (6,59.48)
        (8,59.73) 
    };
    \node at (axis cs:2,59.86) [anchor=south, font=\footnotesize, fill=white] {59.86};

    \addplot[
        mark=square*,
        thick,
        orange
    ] coordinates {
        (1,59.86)
        (2,60.92)
        (4,61.04)
        (6,61.19)
        (8,60.85)
    };
    \node at (axis cs:6,61.19) [anchor=north, font=\footnotesize, fill=white] {61.19};

    \end{axis}
    \end{tikzpicture}
    
    \caption{Accuracy sensitivity according to loss weight on the CIFAR-100 (Top) and TinyImageNet (Bottom).}
    \label{fig:figkd_tuning}
\end{figure}

We also investigate the sensitivity of FiGKD to the loss weights $\alpha$ (cross-entropy loss) and $\beta$ (high-frequency detail loss), as shown in Fig.~\ref{fig:figkd_tuning}. All experiments use the ResNet32x4–ResNet8x4 teacher–student pair on CIFAR-100 and TinyImageNet. The blue line shows the results when the CE loss weight is fixed to 1 and the high-frequency loss weight is varied. The orange line represents the results when the high-frequency loss weight is fixed to 2 and the CE loss weight is varied.

On CIFAR-100, the best accuracy is achieved with $\alpha=2$ and $\beta=2$. For consistency and simplicity, we adopt this setting as the default in all main experiments. However, a closer look at TinyImageNet reveals that the optimal setting may vary by dataset: the best result is observed with $\alpha=6$ and $\beta=2$. This suggests that the best-performing hyperparameters for one dataset may not necessarily be optimal for another, and that further gains could be possible by tuning $\alpha$ and $\beta$ specifically for each dataset. In other words, the current reported results do not represent the upper bound of FiGKD's potential performance; there is still room for improvement with dataset-specific tuning.

It is worth noting that, even without extensive hyperparameter search, FiGKD consistently outperforms the strong MLKD baseline on both datasets. On CIFAR-100, all tested values of $\alpha$ (with $\beta=2$) exceed the MLKD baseline (77.08\%). Likewise, on TinyImageNet, combinations with $\beta=2$ and $\alpha \in \{2, 4, 6, 8\}$ outperform the MLKD baseline (60.67\%). These findings highlight both the robustness of FiGKD and the opportunity for further improvement through careful hyperparameter optimization.

\begin{table}[tbp]
\caption{
{Ablation study on the decomposition level $J$.}
}
\centering
\resizebox{0.85\columnwidth}{!}{%
\begin{tabular}{c|cccc}
\toprule
\multirow{2}{*}{$J$} & ResNet32x4 & VGG13 & WRN-40-2 & VGG13 \\
& ResNet8x4 & VGG8 & WRN-16-2 & MobileNet-V2 \\

\midrule
\textbf{1} & \textbf{78.05} & \textbf{76.17} & \textbf{77.06} & \textbf{71.81} \\
2 & 78.05 & 75.93 & 76.80 & 71.25 \\
3 & 77.97 & 75.77 & 76.52 & 71.11 \\
\bottomrule
\end{tabular}
}
\label{tab:ablation_J}
\end{table}

\subsubsection{{Decomposition Level}}
\label{subsubsec:level}

We investigate the impact of the decomposition level $J$ on the distillation performance. As shown in Table~\ref{tab:ablation_J}, we compared the default setting ($J=1$) with deeper decomposition levels ($J=2$ and $J=3$) across four diverse teacher-student pairs on CIFAR-100, including both homogeneous and heterogeneous architectures. The results demonstrate that increasing the decomposition depth does not improve performance, while $J=1$ consistently achieves the highest accuracy across all architectures.

This behavior is explained by the structural constraints of the logit map. In CIFAR-100, the reshaped logit map has a small spatial resolution of $10 \times 10$. 
A single-level DWT ($J=1$) effectively isolates high-frequency semantic details at a resolution of $5 \times 5$. 
Further decompositions reduce the spatial support to $3 \times 3$ ($J=2$) and $2 \times 2$ ($J=3$). 
Given the already compact size of the logit map, these deeper levels fail to provide additional discriminative information compared to the first level, while introducing unnecessary computational overhead. 
Therefore, we adopt $J=1$ as the default setting, as it effectively preserves essential structural fidelity without the complexity of deeper decomposition.

\begin{table}[tbp]
\caption{
{Impact of Gaussian noise injection into high-frequency components.}
}
\centering
\resizebox{1.0\linewidth}{!}{%
\begin{tabular}{c|cccc}
\toprule
\multirow{2}{*}{Noise Std ($\sigma$)} & ResNet32x4 & VGG13 & WRN-40-2 & WRN-40-2 \\
& ResNet8x4 & VGG8 & WRN-16-2 & ShuffleNet-V1 \\
\midrule
\textbf{0.0} & \textbf{78.05} & \textbf{76.17} & \textbf{77.06} & \textbf{78.84} \\
0.05 & 77.94 & 75.90 & 77.05 & 78.67 \\
0.10 & 77.89 & 76.01 & 76.85 & 78.45 \\
0.20 & 77.84 & 76.07 & 76.88 & 78.31 \\
0.50 & 77.71 & 75.94 & 76.93 & 78.51 \\
\bottomrule
\end{tabular}
}
\label{tab:noise_robustness}
\end{table}

\subsubsection{{Robustness to Noise Injection}}

\begin{figure*}[htbp]
  \centering
  \includegraphics[width=0.8\linewidth]{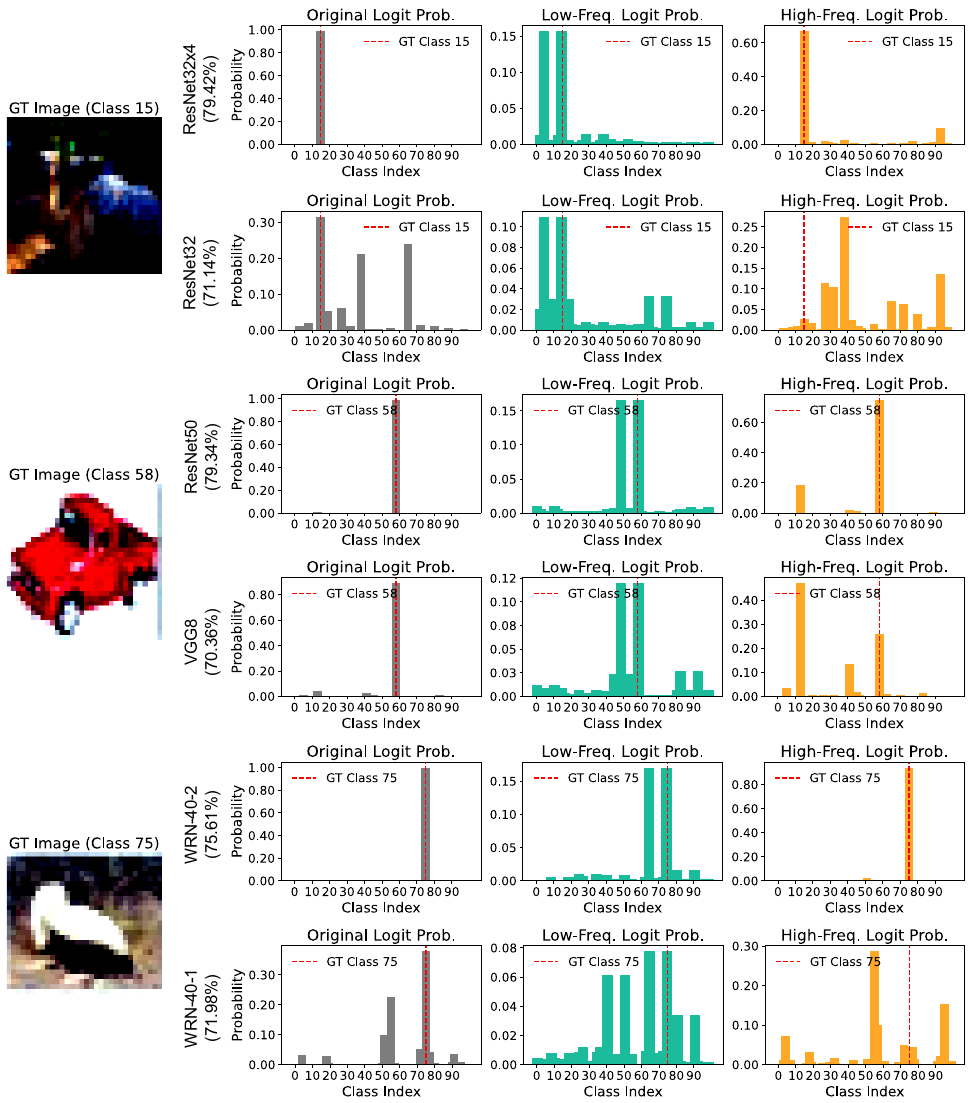}
  \caption{Visualization of probability distributions across frequency components for different model pairs on CIFAR-100.}
  \label{fig:various_logit_analysis}
\end{figure*}

We investigate the robustness of FiGKD against  noise in the high-frequency logit components. 
To this end, we conducted an experiment in which Gaussian noise $\mathcal{N}(0, \sigma^2)$ was injected into the high-frequency subbands of the teacher’s logits during the distillation process. 
To ensure generality, we evaluated four representative teacher-student pairs on CIFAR-100, including both homogeneous and heterogeneous architectures. The standard deviation $\sigma$ of the injected noise varied from 0.0 to 0.5.

As presented in Table~\ref{tab:noise_robustness}, the performance of FiGKD remains remarkably stable, fluctuating only marginally even under high noise levels ($\sigma=0.5$). 
This indicates that FiGKD is robust to perturbations in high-frequency components. 
We attribute this to the dominance of fine-grained semantic patterns. The high-frequency components extracted from the teacher encode meaningful inter-class variations rather than unstructured random noise. 
Because these structural patterns are consistent and semantically distinct, they remain recognizable to the student even under perturbation, ensuring that the essential discriminative cues are effectively transferred.

\subsubsection{{Impact of Dimension}}

\begin{table*}[htbp]
\caption{Comparison between 1D-DWT (flattened logits) and 2D-DWT (reshaped logits) on CIFAR-100. SV1/SV2 denote ShuffleNet-V1/V2, and MV2 denotes MobileNet-V2.}
\label{tab:1d_vs_2d}
\centering
\resizebox{0.85\linewidth}{!}{%
\begin{tabular}{c|cccc|cccc|c}
\toprule
{Teacher} & ResNet32x4 & WRN-40-2 & WRN-40-2 & VGG13 & WRN-40-2 & ResNet32x4 & ResNet32x4 & VGG13 & \multirow{2}{*}{{Avg.}} \\
{Student} & ResNet8x4 & WRN-16-2 & WRN-40-1 & VGG8 & SV1 & SV1 & SV2 & MV2 & \\
\midrule
1D-DWT & 77.75 & 76.92 & 75.61 & 75.42 & 78.39 & 78.11 & 79.03 & 71.15 & 76.55 \\
\textbf{2D-DWT} & \textbf{78.05} & \textbf{77.06} & \textbf{76.04} & \textbf{76.17} & \textbf{78.84} & \textbf{78.70} & \textbf{79.64} & \textbf{71.81} & \textbf{77.04} \\
\midrule
\textbf{$\Delta$} & \textbf{+0.30} & \textbf{+0.14} & \textbf{+0.43} & \textbf{+0.75} & \textbf{+0.45} & \textbf{+0.59} & \textbf{+0.61} & \textbf{+0.66} & \textbf{+0.49} \\
\bottomrule
\end{tabular}
}
\end{table*}

\begin{figure*}[tbp]
  \centering
  \includegraphics[width=0.65\linewidth]{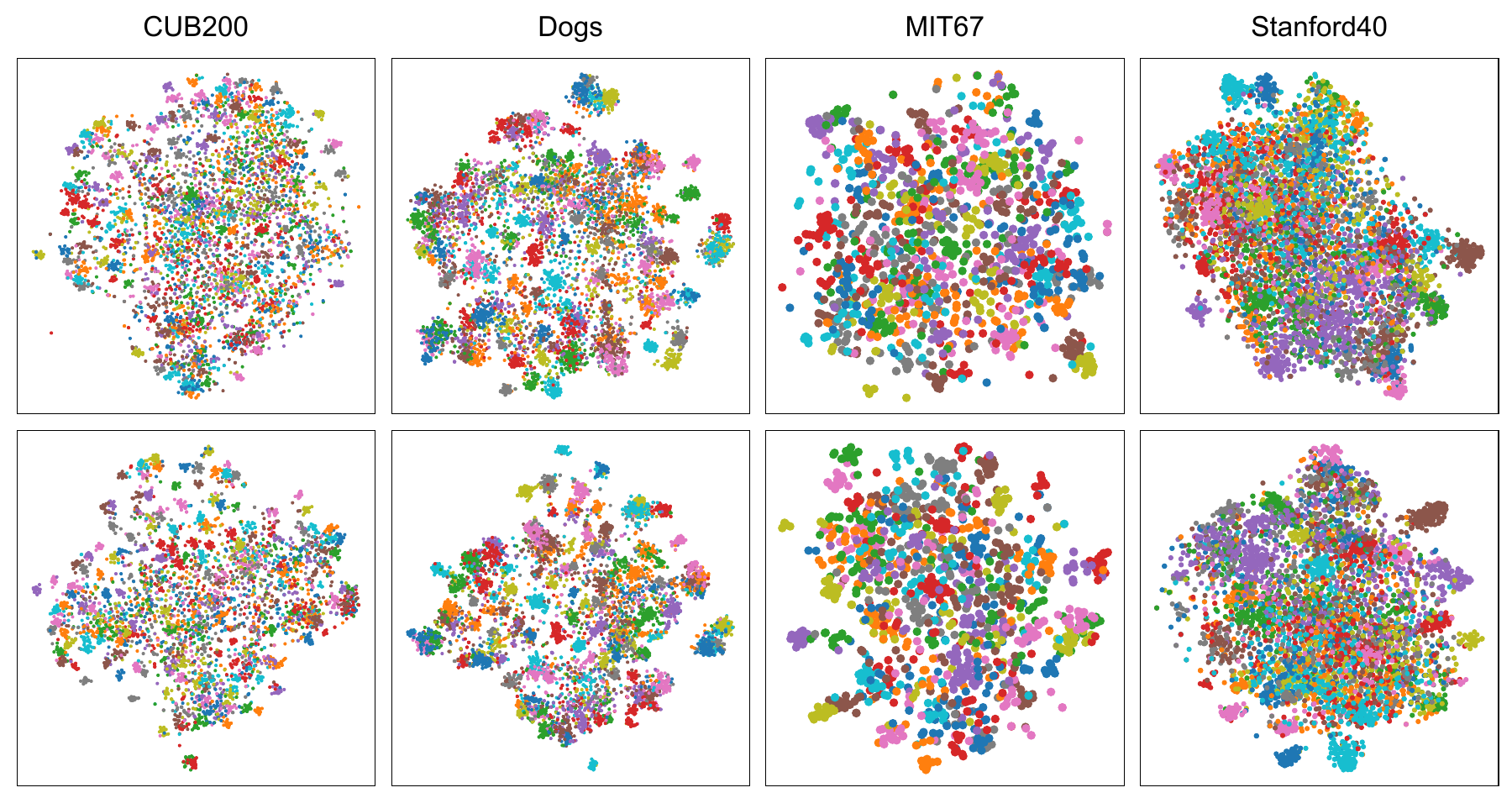}
\caption{t-SNE visualization.}
  \label{fig:tsne}
\end{figure*}

To empirically verify the effectiveness of the proposed 2D transformation, we compared our method with a variant that applies 1D-DWT on the original flattened logits ($1 \times C$). In the 1D setting, the wavelet transform is applied along the single class dimension.

Table~\ref{tab:1d_vs_2d} summarizes the results across eight teacher-student pairs, including both homogeneous and heterogeneous architectures. The proposed 2D-DWT consistently outperforms the 1D-DWT baseline in all cases, yielding an average accuracy improvement of +0.49\%.

We attribute this performance gain to the multi-view supervision capability of the 2D-DWT. By decomposing logits into three orthogonal subbands (horizontal, vertical, and diagonal), the 2D approach enforces structural constraints along multiple independent axes, whereas 1D-DWT is limited to a single frequency axis.

\subsection{Visualization}
\label{subsec:visualizations}

\subsubsection{Frequency Logit Visualization}

To gain deeper insight into how model capacity relates to the frequency composition of logits, we present visualizations in Fig.~\ref{fig:various_logit_analysis}. For each representative sample, the model’s predicted logits are decomposed using 2D DWT into two components—low-frequency and high-frequency—which are then compared against the original logits. The visualization is arranged such that each row corresponds to a specific model and input sample, with the original logits shown in the left column, the low-frequency component in the middle, and the high-frequency component on the right.

The results reveal a striking qualitative difference between high-capacity and low-capacity models. For high-capacity models, the high-frequency logit component alone is often sufficient to correctly predict the target class—the peak corresponding to the true class remains clearly visible even after removing low-frequency information. In contrast, for low-capacity models, the high-frequency logit component no longer distinctly highlights the target class: the distribution appears more diffuse, and the location of the true class is less apparent. This indicates that compact models have difficulty encoding precise class-discriminative information in the high-frequency domain.

These qualitative findings reinforce the core motivation behind FiGKD: by distilling high-frequency knowledge from a stronger teacher, we enable the student to capture subtle semantic cues that would otherwise be difficult to learn—particularly in fine-grained recognition scenarios.

\subsubsection{tSNE}

To visually assess the representational capacity of the proposed distillation method, we employ t-SNE to project feature embeddings into a two-dimensional space. Fig.~\ref{fig:tsne} displays the 2D projections of features extracted from student models trained on four benchmark datasets: CUB200, Dogs, MIT67, and Stanford40. In all cases, the teacher model is ResNet34 and the student is ResNet18. The top row shows student representations obtained via standard knowledge distillation (KD), while the bottom row presents embeddings learned using FiGKD.

Across all datasets, FiGKD consistently produces more semantically meaningful feature spaces than standard KD, characterized by clearer inter-class separability and tighter intra-class clustering. This improvement is particularly evident in challenging datasets such as Dogs, where FiGKD produces more structured and discriminative feature clusters—indicating stronger representational quality for fine-grained and visually diverse recognition tasks.

These results demonstrate that FiGKD effectively transfers high-frequency, semantically rich knowledge from teacher to student, enhancing both quantitative performance and the qualitative structure of learned feature representations. The t-SNE visualizations further support that FiGKD substantially improves the discriminability of deep features compared to conventional KD.

\subsection{{Training Overhead}}
\label{subsec:overhead}

\begin{table}[tbp]
\centering
\caption{Comparison of training overhead with different decomposition levels ($J$) on CIFAR-100.}
\label{tab:training_cost}
\resizebox{0.7\linewidth}{!}{%
\begin{tabular}{c|cc}
\toprule
Method & {Time (s/batch)} & {Overhead (\%)} \\
\midrule
KD (Baseline) & 0.1040 & - \\
\midrule
\textbf{FiGKD ($J=1$)} & \textbf{0.1061} & \textbf{+2.0} \\
FiGKD ($J=2$) & 0.1243 & +19.5 \\
FiGKD ($J=3$) & 0.1353 & +30.1 \\
\bottomrule
\end{tabular}
}
\end{table}

\begin{table*}[tbp]
\centering
\caption{Comparison of model complexity on TinyImageNet.}
\label{tab:model_complexity}
\resizebox{0.7\linewidth}{!}{%
\begin{tabular}{c|cccccc}
\toprule
\multirow{2.5}{*}{Model Pairs}          & \multicolumn{2}{c}{Teacher} & \multicolumn{2}{c}{Student} & \multicolumn{2}{c}{Ratio} \\ \cmidrule{2-7} 
                                      & Params (M)    & FLOPs (M)   & Params (M)    & FLOPs (M)   & Params       & FLOPs      \\ \midrule
R110 $\rightarrow$ R32         & 1.74  & 1029.58 & 0.48 & 281.56 & 3.63× & 3.66× \\
R32x4 $\rightarrow$ R8x4       & 7.46  & 4352.82 & 1.26 & 714.26 & 5.92× & 6.09× \\
WR402 $\rightarrow$ WR162      & 2.27  & 1322.61 & 0.72 & 409.30 & 3.15× & 3.23× \\
WR402 $\rightarrow$ WR401      & 2.27  & 1322.61 & 0.58 & 339.36 & 3.91× & 3.90× \\
V13 $\rightarrow$ V8           & 9.51  & 917.16  & 4.02 & 273.43 & 2.37× & 3.35× \\ \midrule
WR402 $\rightarrow$ SV1        & 2.27  & 1322.61 & 1.05 & 42.88  & 2.16× & 30.84× \\
R50 $\rightarrow$ MV2          & 23.91 & 5246.70 & 0.94 & 29.23  & 25.44× & 179.50× \\
R32x4 $\rightarrow$ SV1        & 7.46  & 4352.82 & 1.05 & 42.88  & 7.10× & 101.51× \\
R32x4 $\rightarrow$ SV2        & 7.46  & 4352.82 & 1.46 & 186.82 & 5.11× & 23.30× \\
V13 $\rightarrow$ MV2          & 9.51  & 917.16  & 0.94 & 29.23  & 10.12× & 31.38× \\
\bottomrule
\end{tabular}
}
\end{table*}

To verify the efficiency of FiGKD, we analyzed the computational overhead introduced by the logit-based wavelet decomposition. 
Table~\ref{tab:training_cost} presents the training time per batch using the representative pair ResNet32x4-ResNet8x4 on the CIFAR-100 dataset, conducted on an NVIDIA RTX 3090 GPU.

Compared to the standard KD baseline ($0.1040$ s/batch), FiGKD with the default setting ($J=1$) requires $0.1061$ s/batch. 
The marginal increase of approximately 2.1 ms ($\sim$2.0\%) confirms that the additional operations—logit reshaping, forward DWT, and detail loss calculation—incur negligible overhead and do not hinder training efficiency.

We further investigated the effect of the decomposition level $J$ on training efficiency. 
As shown in Table~\ref{tab:training_cost}, increasing $J$ from 1 to 3 leads to a substantial increase in computational cost ($+19.5\%$ for $J=2$ and $+30.1\%$ for $J=3$). 
This overhead arises because deeper decompositions require repeated DWT operations over progressively smaller subbands, accumulating additional transform and memory access costs.
%
%
Given that $J=1$ achieves the best accuracy (Sec.~\ref{subsubsec:level}) while maintaining the lowest training overhead, we conclude that our single-level decomposition strategy is the optimal choice considering both performance and efficiency.

\subsection{Teacher–Student Scale Comparison}
\label{subsec:efficiency}

Table~\ref{tab:model_complexity} summarizes the number of parameters and FLOPs for the teacher and student models used in our experiments on TinyImageNet. The last two columns show the teacher-to-student ratios, indicating how many times larger the teacher model is compared to the student in terms of both parameter count and computational cost. 

As shown, teacher models are consistently several times larger than their student counterparts in both parameter count and computational cost. Notably, the ResNet50-MobileNetV2 pair represents an extreme case, with the teacher model being over 25 times larger in parameters and nearly 180 times more expensive in FLOPs.

Despite substantial differences in model size, FiGKD enables student models to effectively absorb fine-grained knowledge from their respective teachers, significantly narrowing the performance gap while maintaining high computational efficiency. These results underscore the robustness and applicability of our method across a wide range of model sizes, particularly in highly compressed deployment scenarios.

\section{{Conclusion}}
\label{sec:conclusion}

In this paper, we presented FiGKD (Fine-Grained Knowledge Distillation), a novel logit-based distillation framework that introduces a frequency-domain perspective to knowledge transfer. By applying the Discrete Wavelet Transform (DWT) to the teacher’s output logits, FiGKD decomposes them into low- and high-frequency components and selectively distills only the high-frequency information, which captures subtle, fine-grained semantic relationships. This targeted distillation enables compact student models to better replicate the nuanced decision boundaries learned by high-capacity teacher networks. Unlike conventional KD approaches that either treat all logit components uniformly or require access to high-dimensional feature representations, FiGKD is simple, architecture-agnostic, and easy to implement. Our frequency-aware strategy yields consistent performance gains on standard classification benchmarks such as CIFAR-100 and TinyImageNet, and shows especially strong improvements on fine-grained visual recognition (FGVR) tasks, where small semantic differences are critical. Comprehensive experiments and ablation studies demonstrate the effectiveness and robustness of FiGKD across various teacher–student pairs and hyperparameter settings. We further provide visual and structural analyses to validate the role of high-frequency components in capturing class-discriminative signals. Overall, FiGKD opens up a new direction for logit-based knowledge distillation by bridging semantic signal processing with model compression. We believe this frequency-domain perspective offers promising avenues for future research, including adaptive frequency selection and extensions to other modalities such as vision–language models.

\section*{Acknowledgments}
Special thanks to Young Hwa Sung and Chong Hui Kim for providing an excellent research environment. This work was supported by the Korean Government.

\appendix
\section{{Impact of Different Wavelet Bases}}
\label{supp:wavelet_bases}

\vspace{-0.5cm}

\begin{table}[h]
\centering
\caption{Comparison of FiGKD performance using different wavelet bases (Haar, db2, bior1.3) across various teacher-student pairs on CIFAR-100.}
\resizebox{1.0\linewidth}{!}{
\begin{tabular}{c|cccccc}
\toprule
\multirow{2}{*}{Wavelet} & \multicolumn{2}{c}{ResNet32x4} & \multicolumn{2}{c}{VGG13} & \multicolumn{2}{c}{WRN-40-2} \\
& ResNet8x4    & SV2   & VGG8     & MV2   & WRN-16-2   & SV1   \\                          
\midrule
Haar                     & 78.05        & 79.64           & 76.17    & 71.81          & 77.06      & 78.84           \\
db2                      & 77.80        & 79.09           & 75.90    & 71.29          & 76.92      & 78.71           \\
bior1.3                  & 77.39        & 79.11           & 75.37    & 70.90          & 76.78      & 77.96           \\ 
\bottomrule
\end{tabular}}
\label{tab:wavelet_bases}
\end{table}

We explore the impact of different wavelet bases on the distillation performance of FiGKD.
While the Haar wavelet is adopted as the default choice in the main experiments due to its simplicity and computational efficiency, we further investigate whether FiGKD remains effective with other wavelet families.
Specifically, we compare the Haar wavelet with two representative alternatives: Daubechies-2 (db2), which features smoother basis functions, and Biorthogonal 1.3 (bior1.3), which is symmetric.
The experiments were conducted on CIFAR-100 across diverse teacher–student pairs using the same hyperparameter settings optimized for Haar.
The results are summarized in Table~\ref{tab:wavelet_bases}.

As shown in Table~\ref{tab:wavelet_bases}, the Haar wavelet achieves the highest accuracy.
This is expected, as the current hyperparameters were tuned specifically for the Haar basis.
Nevertheless, db2 and bior1.3 also yield competitive performance with only marginal degradation.
This suggests that FiGKD is not strictly dependent on a specific wavelet basis and remains generally applicable across different wavelet families, although alternative bases may benefit from dedicated hyperparameter tuning.

  \bibliographystyle{elsarticle-num} 
  \bibliography{references}
  
\end{document}